\def\year{2023}\relax
\documentclass[letterpaper]{article} 
\usepackage{aaai23}  
\usepackage{times}  
\usepackage{helvet}  
\usepackage{courier}  
\usepackage[hyphens]{url}  
\usepackage{graphicx} 
\usepackage{natbib}  
\usepackage{caption} 
\DeclareCaptionStyle{ruled}{labelfont=normalfont,labelsep=colon,strut=off} 
\usepackage{algorithm}
\usepackage{algorithmic}

\usepackage{newfloat}
\usepackage{listings}
\floatstyle{ruled}
\newfloat{listing}{tb}{lst}{}
\floatname{listing}{Listing}
\title{FedSSO: A Federated Server-Side Second-Order Optimization Algorithm}
\author {
    Xin Ma\textsuperscript{\rm 1} 
    Renyi Bao\textsuperscript{\rm 1} 
    Jinpeng Jiang\textsuperscript{\rm 1*} 
    Yang Liu\textsuperscript{\rm 2\dag} 
    Arthur Jiang\textsuperscript{\rm 1} 
    Jun Yan\textsuperscript{\rm 1} 
    Xin Liu\textsuperscript{\rm 3} 
    Zhisong Pan\textsuperscript{\rm 3}
}
\affiliations {
    \textsuperscript{\rm 1} Yidu Cloud Technology Inc., Beijing, China\\
    \textsuperscript{\rm 2} Institute for AI Industry Research, Tsinghua University, Beijing, China\\
    \textsuperscript{\rm 3} Army Engineering University of PLA, China\\
    xin.ma0206@gmail.com, renyi.bao@yiducloud.cn, jinpeng.jiang@yiducloud.cn, liuy03@air.tsinghua.edu.cn, arthursjiang@gmail.com, jun.yan@yiducloud.cn, liuxin@aeu.edu.cn,
    hotpzs@hotmail.com
}

\usepackage{bibentry}
\usepackage{tabularx}

\usepackage[utf8]{inputenc} 
\usepackage[T1]{fontenc}    
\usepackage{hyperref}       
\usepackage{url}            
\usepackage{booktabs}       
\usepackage{amsfonts}       
\usepackage{nicefrac}       
\usepackage{microtype}      
\usepackage{xcolor}         

\usepackage{comment}
\usepackage{algorithm}
\usepackage{algorithmic}
\usepackage{multicol}
\usepackage{graphicx}
\usepackage{titletoc}

\usepackage{caption,subcaption}
\usepackage{amsmath}
\usepackage{amssymb}
\usepackage{mathtools}
\usepackage{amsthm}
\usepackage{makecell}

\usepackage[capitalize,noabbrev]{cleveref}

\theoremstyle{plain}

\theoremstyle{definition}

\theoremstyle{remark}

\usepackage[flushleft]{threeparttable}
\usepackage{booktabs,amsmath,siunitx}

\usepackage{minitoc}

\begin{document}
\maketitle
\begin{abstract}
In this work, we propose FedSSO, a server-side second-order optimization method for federated learning (FL). In contrast to previous works in this direction,  we employ a server-side approximation for the Quasi-Newton method without requiring any training data from the clients. In this way, we not only shift the computation burden from clients to server, but also eliminate the additional communication for second-order updates between clients and server entirely.  We provide theoretical guarantee for convergence of our novel method, and empirically demonstrate our fast convergence and communication savings in both convex and non-convex settings.
\end{abstract}

\section{Introduction}

{F}{ederated}
\cite{1975Some}
 Learning (FL) facilitates the practical applications of machine learning techniques in cross-silo scenarios by collaboratively training the distributed private data while preserving users’ privacy~\cite{DBLP:journals/tist/YangLCT19}, which is especially important in privacy-sensitive domains like finance and healthcare. Depending on how data is partitioned, FL can be further categorized into horizontal FL ~\cite{DBLP:conf/aistats/McMahanMRHA17} and vertical FL ~\cite{DBLP:journals/corr/abs-1901-08755}.  In a typical FL process like FedAvg~\cite{DBLP:conf/aistats/McMahanMRHA17}, clients perform multiple rounds of local gradient updates and send their updates to a server, who will then perform global aggregation before sending the global updates back to clients for next iteration. However, FedAvg only works well for IID scenarios ~\cite{DBLP:conf/mlsys/LiSZSTS20}. The cross-silo nature of FL also introduces nontrivial challenges. Take the healthcare domain as an example. Distributions of disease and patients from different hospitals are often not independent and identically distributed (Non-IID), which may degrade the convergence and performance of the trained global model~\cite{DBLP:journals/corr/abs-1806-00582}. Furthermore, the computation capability and network stability of each hospital are limited and diverse. 

Over the years, many algorithms have been proposed to address the Non-IID issue, such as FedProx ~\cite{DBLP:conf/mlsys/LiSZSTS20} and Scaffold ~\cite{DBLP:conf/icml/KarimireddyKMRS20}. These improvements focus on first-order optimization, but incur high iteration and communication cost ~\cite{9443421}. 
Recently second-order Newton-type optimization strategies, such as  FedDANE~\cite{DBLP:conf/icml/ZhangL15a} and FedNL ~\cite{DBLP:journals/corr/abs-2106-02969} are proposed to further improve the model convergence in Non-IID scenarios. However, the implementation of classic Newton-type methods on the clients of FL is not admittedly efficient, due to the frequent communication of gradients as well as second-order updates, such as Hessians.

In this paper, we propose a \textbf{fed}erated \textbf{s}erver-side \textbf{s}econd-order \textbf{o}ptimization algorithm, FedSSO, attempting to address the massive communication overload issue with convergence guarantee. Our framework adopts a similar workflow as FedAvg but applies a Quasi-Newton method to generate an approximately global Hessian matrix on the server-side. Specifically, FedSSO will first perform multiple local upstate on the client-side just like FedAvg, and the approximated global gradient will be calculated by the aggregation of gradients on the server. Then, based on the global gradient, we approximate the global Hessian matrix by the Quasi-Newton method. Finally, Quasi-Newton descent will be performed on the server-side and the updated global model is sent to clients. We provide theoretical proof and analysis on the convergence properties of FedSSO. To the best of our knowledge, FedSSO is the first approach which applies the Quasi-Newton optimization method on the server-side to reduce communication overloads with convergence guaranteed in FL. Furthermore, through extensive experiments, FedSSO has shown its advantage compared to its counterparts, on both convex and non-convex settings with different Non-IID distribution.

\begin{figure}[h]
    \centering
    \includegraphics[width=1.05 \linewidth]{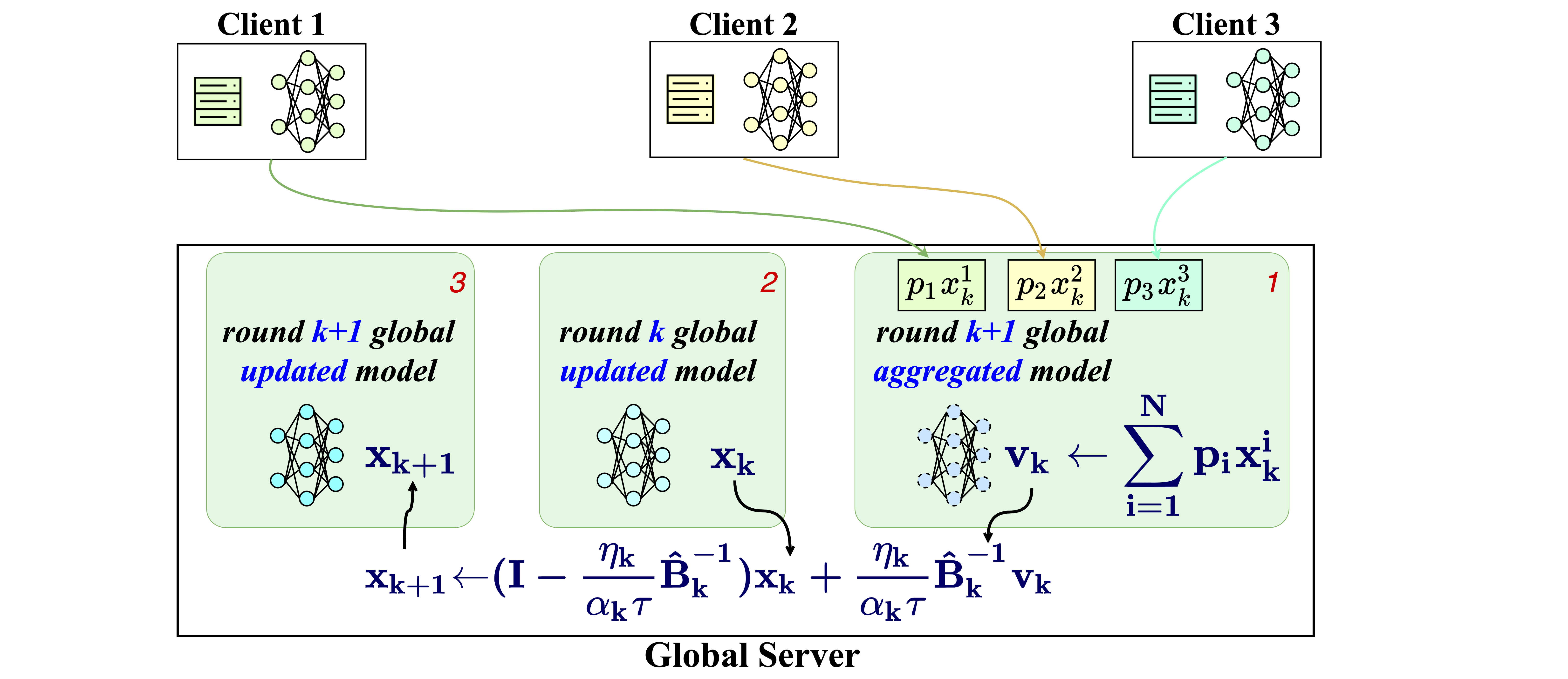}
    \caption{Overview of the FedSSO workflow.}
    \label{fig:fedsso_overall}
\end{figure}


In summary, the main contributions of this paper are as follows:
\begin{itemize}
    \item We propose a novel server-side second-order optimization method in FL, which effectively calculates the global gradients and Hessians in a centralized manner and completely eliminates the communication of second-order Hessian terms between clients and server as compared to existing second-order optimization methods, achieving significant improvement in overall communication efficiency. 
    \item We conduct theoretical analysis on the convergence of our method and prove that FedSSO reaches a convergence rate of $\mathcal{O}(\frac{1}{k})$, which is comparable to Stochastic Quasi-Newton method.
    \item Furthermore, because FedSSO shifts the computation burden of second-order updates from the clients to the server, no additional computations are required on the clients, making it more applicable for FL scenarios with resource constrained edges.
\end{itemize}

\section{Related Work}
\subsection{Federated Learning with Non-IID data.}FedAvg is one of the most common algorithms in FL~\cite{DBLP:conf/aistats/McMahanMRHA17}. However, FedAvg can not adequately address the convergence and communication efficiency issues caused by the Non-IID data~\cite{DBLP:conf/mlsys/LiSZSTS20,li2021federated,zhang2021fedpd,xu2021fedcm}. For example, 
~\cite{DBLP:conf/iclr/LiHYWZ20} describes a trade-off between convergence rate and communication, and points that data heterogeneity can negatively impact the convergence. 
Scaffold proves that unstable convergence of FedAvg results from `client-drift' phenomenon when data is Non-IID~\cite{DBLP:conf/icml/KarimireddyKMRS20}.

\subsection{First-order federated optimization methods.}
The main idea of first-order optimization methods 
is to reduce variance inherent in the process of gradient estimation. Stochastic algorithms form the basis of this category of methods, such as Stochastic Average Gradient (SAG)~\cite{roux2012stochastic} and Stochastic Variance Reduced Gradient (SVRG)~\cite{johnson2013accelerating}. 
The first-order optimization ideas to reduce variance are widely applied to FL. 
FedSGD~\cite{2016FedSGD} is a centralized SGD method applied to FL, which is equivalent to FedAvg with only one local step. Meantime, FedAvg can use multiple local upstate to reduce communication cost and accelerate convergence.
Scaffold~\cite{DBLP:conf/icml/KarimireddyKMRS20} tries to estimate the update directions for server model and each client, which are used to estimate the degree of client-draft and correct the local updates. HarmoFL~\cite{DBLP:journals/corr/abs-2112-10775} tries to mitigate the drift problem from both the client and server sides. FedProx~\cite{DBLP:conf/mlsys/LiSZSTS20} utilizes a strategy of adding a proximal term with the subproblem on each client to improve the stability. FedAC~\cite{DBLP:conf/nips/YuanM20} transforms stand-alone Nesterov Momentum into parallel federated optimization algorithm, but it has more hyperparameters and needs more communication load. STEM ~\cite{DBLP:conf/nips/KhanduriSYHLRV21}reduces communication overhead, but does not solve the client-drift problem. FedNova~\cite{DBLP:conf/nips/WangLLJP20} averages client gradients from different number of local updates. LD-SGD incorporates arbitrary update schemes that alternate between multiple local updates and multiple Decentralized SGDs ~\cite{DBLP:journals/corr/abs-1910-09126}.

In addition to the above first-order optimization methods which all focus on training a global model, another group of methods focus on training customized model on clients, i.e., personalized federated learning
~\cite{DBLP:conf/nips/SmithCST17,DBLP:conf/nips/0001MO20,DBLP:journals/corr/abs-1912-00818,DBLP:conf/aaai/HuangCZWLPZ21,dinh2020personalized,zhang2020personalized}. M{\scriptsize{OCHA}}~\cite{DBLP:conf/nips/SmithCST17} is developed as a multi-task learning scheme to fit separate weight vectors to the data in each task. FedPer~\cite{DBLP:journals/corr/abs-1912-00818} adds personalization layers after the base layers of networks and trains personalization layers only on local data with SGD. 
~\cite{DBLP:journals/ftml/KairouzMABBBBCC21} gives a comprehensive analysis and comparison on these first-order methods. 

\subsection{Second-order federated optimization methods.}
Here, we focus on some recent work aiming to design communication-efficient second-order optimization algorithms in distributed machine learning settings, such as DANE~\cite{DBLP:conf/icml/ShamirS014}, AIDE~\cite{DBLP:journals/corr/ReddiKRPS16}, DiSCO~\cite{DBLP:conf/icml/ZhangL15a}, 
DONE~\cite{DBLP:journals/corr/abs-2012-05625} and 
LocalNewton with global line search~\cite{DBLP_journals_corr_abs_2109_02388}. Specifically, both DANE and AIDE are approximate Newton-like methods. DANE can solve a general sub-problem available locally using the implicit local Hessian 
~\cite{DBLP:conf/icml/ShamirS014}. 
AIDE (i.e., an inexact variant of DANE) is proposed to match the communication lower bounds. DiSCO can be considered as an inexact damped Newton method, which uses distributed preconditioned conjugate gradient to compute the inexact Newton steps efficiently. DiSCO demonstrates theoretically that its upper bound on number of communication rounds is less than that of DANE~\cite{DBLP:conf/icml/ZhangL15a}. As an inexact distributed Newton-type method, DANE can effectively approximate the true Newton direction using the Richardson iteration for convex functions, and it has been proved theoretically to have a linear-quadratic convergence rate~\cite{DBLP:journals/corr/abs-2012-05625}. Newton-Learn is another communication-efficient scheme incorporating compression strategies for second-order information~\cite{DBLP:conf/icml/IslamovQR21}. 

In the FL scenario, FedDANE extends inexact DANE algorithm to solve the heterogeneity and low participation problems by approximating the full gradients from some sampled devices~\cite{DBLP:conf/acssc/LiSZSTS19}. Based on Newton-Learn~\cite{DBLP:conf/icml/IslamovQR21}, a family of Federated Newton Learn (FedNL) algorithms is proposed to boost the applications of second-order methods in FL~\cite{DBLP:journals/corr/abs-2106-02969}. Furthermore, as a generalization of FedNL with more aggressive compression, Basis Learn (BL) successfully integrates bidirectional compression with any predefined basis for Hessian in order to further decrease the communication between clients and server~\cite{DBLP:journals/corr/abs-2111-01847}. Although these attempts are forward-looking, FedDANE consumes more communication rounds than FedAvg 
, and the FedNL series of algorithms required more computational costs and more communication overloads.

In summary, these existing second-order approaches rely on the clients to perform the computation of global gradient and global Hessian, which will inevitably lead to high communication cost and resource consumption. In contrast, we take a different approach to perform these estimations on the server side, thereby effectively reducing the overall communication cost.



\section{Preliminaries}
\subsection{Problem Statement}
In FL with cross-silo data, the following optimization objective over distributed clients is commonly considered:
\vspace{-0.1in}
\begin{equation}
    \label{object_func}
    \min_x f(x) = {\sum}_{i=1}^N p_i f_i(x),
\end{equation}

where $x$ is the model needed to be updated, $N$ represents the number of distribute clients, $p_i$ is the weight of $i$-th client, and $\sum_{i=1}^N p_i = 1$. Furthermore, we denote the loss function in clients as $\ell$, and denote the $k$-th sample of client $i$ as $\zeta_k^{(i)}$, and a total of $n_i$ samples on the client $i$. Then the local objective $f_i$ can be defined:
\vspace{-0.1in}
\begin{equation}
   f_i(x) = \frac{1}{n_i}{\sum}_{k=1}^{n_i}\ell(x; \zeta_k^{(i)})
\label{eq:loss local}
\end{equation}

\vspace{-0.1in}

Let $x^{(i)}_{k,j}$ represent the local model of $i$-th client in the $j$-th local epoch of $k$-th round, and $g_i(\cdot)$ represent i-th client's stochastic gradient, 
we provide details of FedAvg in Algorithm~\ref{alg:FedAvg} in Appendix. 
In the FedAvg framework, clients perform local updates as follows:
\vspace{-0.1in}
\begin{equation}
x_{k,\tau}^{(i)} = x_{k,0}^{(i)} - \alpha_k {\sum}_{j=0}^{\tau-1} \nabla f_i(x_{k,j}^{(i)}) 
\label{eq:Fedavg local}
\end{equation}
\vspace{-0.1in}

After local updates, clients send their local updates to the server, where aggregation is performed:
\vspace{-0.1in}
\begin{equation}
x_{k+1} = {\sum}_{i=1}^{N} p_i x_{k,\tau}^{(i)} = x_k - \alpha_k {\sum}_{i=1}^{N} p_i {\sum}_{j=0}^{\tau-1} \nabla f_i(x_{k,j}^{i})
\label{eq:Fedavg server}
\end{equation}
\vspace{-0.1in}

\subsection{BFGS Algorithm}
Newton and Quasi-Newton methods can also effectively solve the unconstrained optimization problems using second-order information. Specifically, Broyden–Fletcher–Goldfarb–Shanno (BFGS) algorithms can be considered as one of the most effective algorithms~\cite{dai2002convergence,DBLP:journals/appml/YuanZZ22}. In BFGS, the expensive of computation of Hessian matrices $H_k$ can be avoided via approximation using $B_k$ (output of BFGS). Specifically, $B_k$ can be updated by Equ.\ref{eq:BFGS}. And Sherman-Morrison formula can directly compute the inversion of $B_i$, which decrease the cost of computing $H_k^{-1}$ greatly~\cite{erway2012limited}.
\begin{equation}
   B_k=B_{k-1}+\frac{y_{k-1}y_{k-1}^\mathrm{ T }}{y_{k-1}^\mathrm{T} s_{k-1}}-\frac{B_{k-1}s_{k-1}s^\mathrm{T}_{k-1}B_{k-1}}{s^\mathrm{T}_{k-1}B_{k-1}s_{k-1}} 
   \label{eq:BFGS}
\end{equation} 
where $y_k=g_{k+1}-g_k$, $s_k = x_{k+1}-x_k$, and $g_k$ is the gradient of epoch $k$.

\section{Methodology}
In this section, we introduce the proposed \textbf{Fed}erated \textbf{S}erver-side \textbf{S}econd-order \textbf{O}ptimization method (FedSSO). 

\subsection{Inspiration}
In FL, the local updates in clients can be used to explore the descent direction for model updates.
On the server side, let $x_k$ denotes the global model at $k$ round, that is, $x_k = \sum_{i=1}^N p_i x_k^{i}$. Equ.\ref{eq:Fedavg server} shows how the global model of the federated process can be updated in FedAvg. It also shows a descent direction. Based on this observation, we try to simulate the gradients needed for the updates of global model by using information of local gradients. Specifically, we consider the optimization at the server and expand the objective function according to the second-order approximation as follows:
\begin{align}
\label{opt_equ}
f(x) \approx & f(x_{k}) + \bigtriangledown f(x_{k})^\mathrm{T}(x - x_{k}) + \dfrac{1}{2}(x - x_{k})^\mathrm{T} H_{k}(x - x_{k})
\end{align}
from which we can obtain the optimal point for the right side as
\begin{equation}
\label{opt_w}
x^* = x_{k} - H_{k}^{-1}\bigtriangledown f(x_{k})
\end{equation}

\begin{figure}[h]
    \centering
    \includegraphics[ width=.95\linewidth]{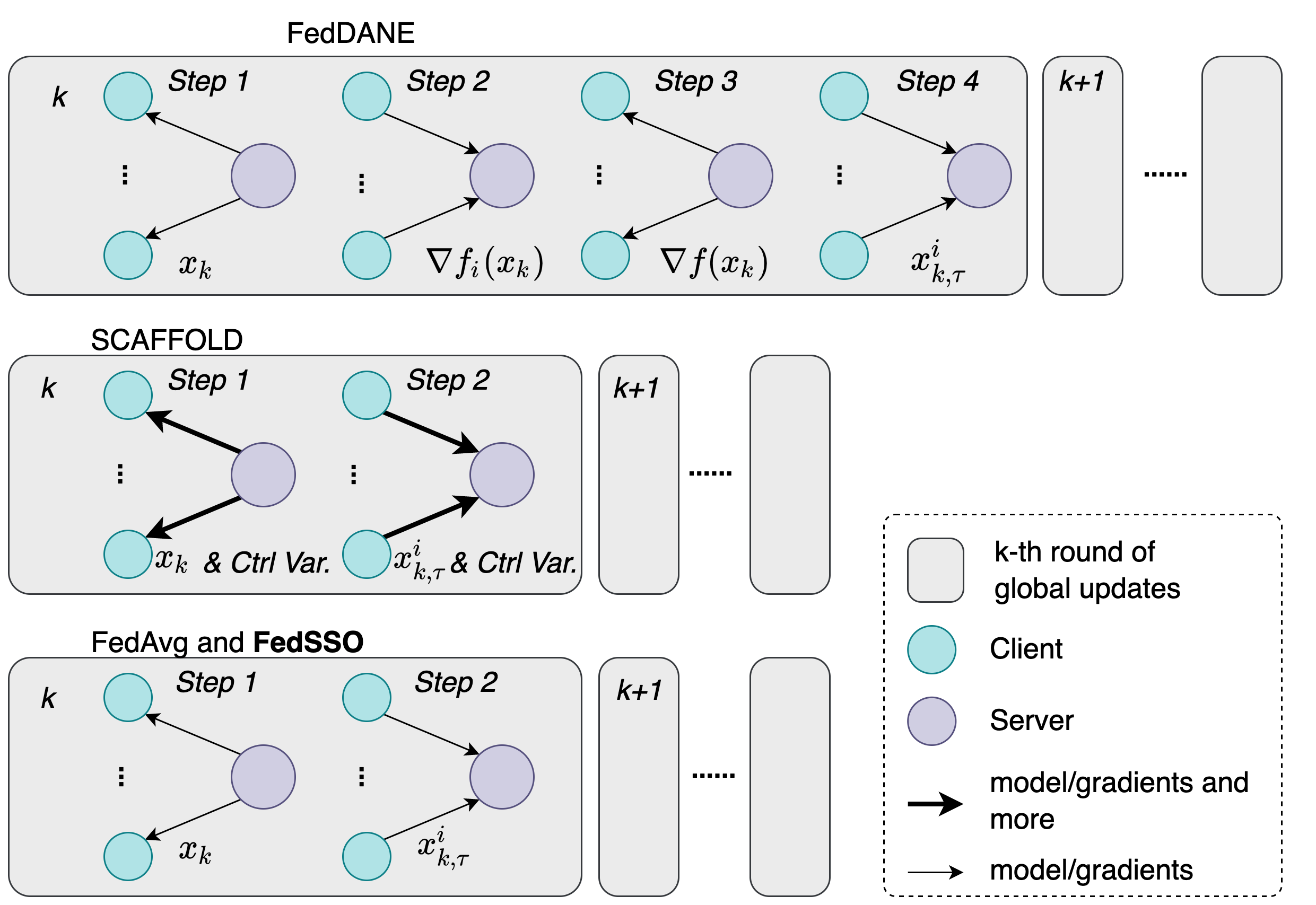}
    \caption{Communication protocols of different federated algorithms: FedDANE, SCAFFOLD, FedAvg and FedSSO.}
    \label{fig:Fed_Algorithms_Diff_on_Com}
\end{figure}

In order to obtain $\nabla f(x_{k})$ and solve Equ.\ref{opt_w}, previous second-order optimization methods such as FedDANE ask clients for their local gradient updates, then send the aggregated gradients back to local clients, who will then solve a local subproblem to obtain a new local model (see Fig.\ref{fig:Fed_Algorithms_Diff_on_Com}). This process requires multiple rounds of communications and local computations. 
In this work, we attempt to use an estimated averaged gradient to replace $\nabla f(x_{k})$ so that only a single round of communication is required between server and clients for each iteration (See Fig. \ref{fig:Fed_Algorithms_Diff_on_Com}). The details of our method are explained in the following sections.

\subsection{Approximation of Global Gradient} 

In order to distinguish from FedAvg, we define the immediate global update for the $k$-th round as $v_k$:
\vspace{-0.04in}
\begin{equation}
\label{v_k}
    v_k = {\sum}_{i=1}^N p_i x_{k,\tau}^{(i)}
\end{equation}
then we approximate the global gradient $\bigtriangledown f(x_{k})$ using the average gradient as:
\begin{equation}
\label{glb_grad}
    \bigtriangledown f(x_k)\approx g(\hat{x}_{k}) = \frac{1}{\alpha \tau}(x_{k} - {v_k})
\end{equation}
where $\hat{x_k}$ denotes a "Lighthouse" point which the average gradients correspond to. 
$\alpha$ and $\tau$ denote the local learning rate and the number of local updates, respectively. We will first prove the existence of $\hat{x_k}$ in Theorem 1.1.

\begin{figure}[htbp]
    \centering
    \includegraphics[width=0.95 \linewidth]{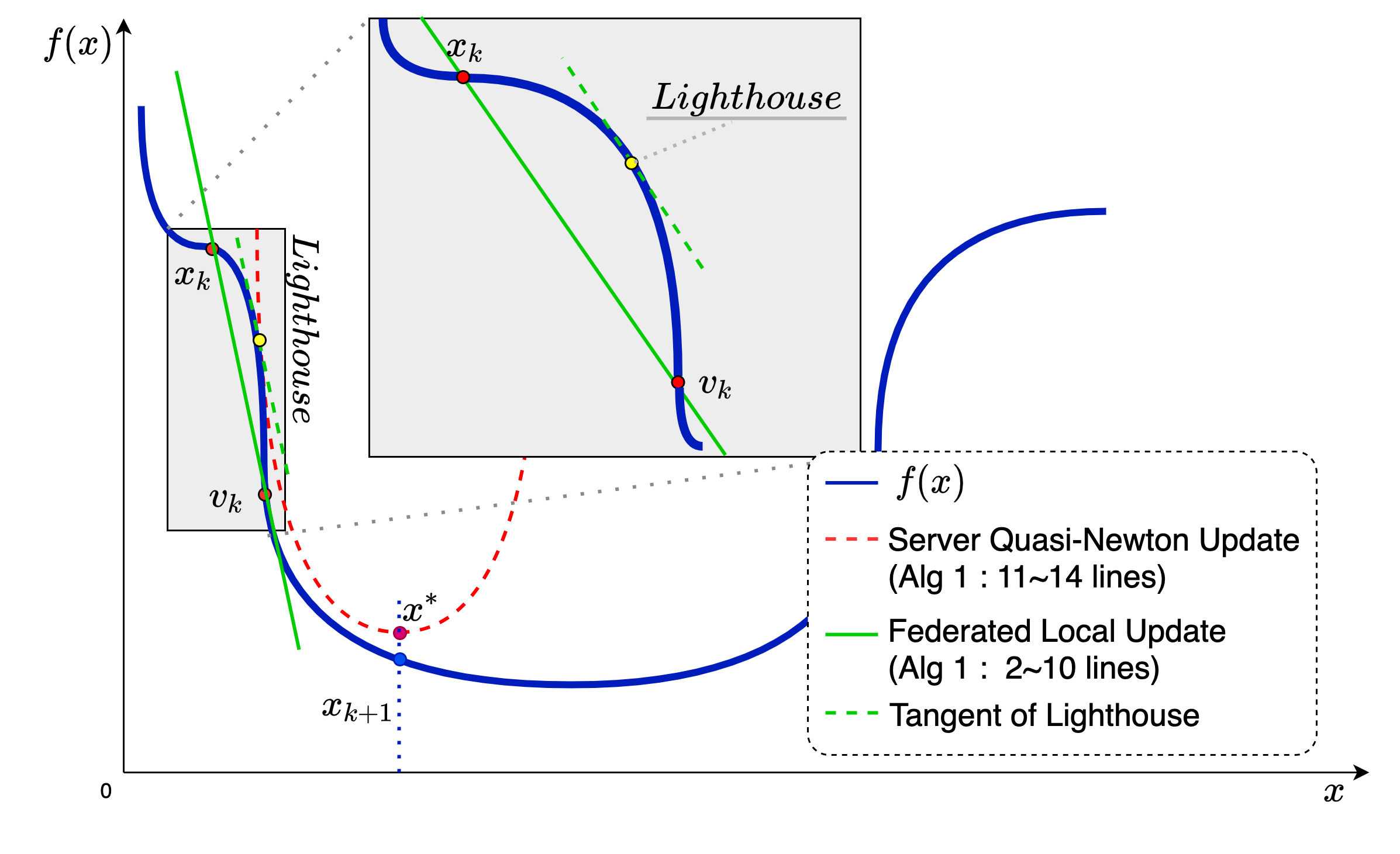}
    \caption{The optimization steps in FedSSO: the path from $x_k$ to $v_k$ represents a federated updating process; The path from ${x_k}$ to $x_{k+1}$ represents server Quasi-Newton process. Lighthouse point is shown.}
    \label{fig:lighthouse}
\end{figure}
Fig\ref{fig:lighthouse} demonstrates the concept of the Lighthouse point. 
If $\hat{x_k}$ exists, it means that we can use the global gradients at $\hat{x_k}$ point to replace the global gradients at $x_k$, and we can substitute $\bigtriangledown f(x_k)$ in Equation~\ref{opt_w}:
\begin{align*}
x_{k+1} &=  x_k - \frac{1}{\alpha \tau} H_{k}^{-1} (x_{k} - {v_k}) \\
        &= (I-\frac{1}{\alpha \tau} H_{k}^{-1})x_{k} + \frac{1}{\alpha \tau} H_{k}^{-1}{v_k}
\end{align*}
Next we will tackle the estimation of the Hessian matrix.
\subsection{Optimization using BFGS algorithm.}
To tackle the high cost for computing and communicating the Hessian matrix, we propose to apply Quasi-Newton method at the server side instead of the client side. This will not only eliminate entirely the communication cost of Hessians, but also alleviate the computation burden on the clients, making the algorithm more favorable and applicable for scenarios with resource constrained edges. 
However since training data locate only on the distributed clients not the server, it is nontrivial to obtain a server-side approximation of Hessians without any training data. In order to tackle this challenge, we first construct a BFGS-like second-order Quasi-Newton approximation with the available gradient approximation $\hat{x_k}$ as:
\begin{equation}
   \hat{B}_k=\hat{B}_{k-1}+\frac{\hat{y}_{k-1} \hat{y}_{k-1}^\mathrm{ T }}{ \hat{y}_{k-1}^\mathrm{T} s_{k-1}}-\frac{B_{k-1}s_{k-1}s^\mathrm{T}_{k-1}B_{k-1}}{s^\mathrm{T}_{k-1}B_{k-1}s_{k-1}} 
   \label{eq:BFGS}
\end{equation} 
where $\hat{y}_{k-1} = g(\hat{x_k})-g(\hat{x}_{k-1})$, $s_{k-1} = x_k - x_{k-1}$.

Note that such an update strategy only requires information available on the server side. Also note that when the server applies BFGS algorithm for approximation, $H_{k} \approx B_k$. As a result, the optimal weights of global model can be computed on server side without sharing the private data as follow:

\vspace{-0.04in}
$$
x_{k+1}
= (I-\dfrac{\eta}{\alpha \tau} \hat{B}_{k}^{-1})x_{k} + \dfrac{\eta}{\alpha \tau} \hat{B}_{k}^{-1}{v_k}
$$
where $\eta$ represents the step length of Quasi-Newton process (See Fig \ref{fig:lighthouse}). The effectiveness of using $\hat{B}$ constructed from $\nabla f(\hat{x_k})$ is validated by our experiments. Here we briefly explain the intuition for adopting such an strategy. In the next section, we perform comprehensive convergence analysis for our proposed method. Our intuition stems from the fact that BFGS is also an approximation to Hessian, and the update strategy contains key information of curvature. Specifically,  In the original BFGS we have  
\begin{align*}
    \frac{y_{k-1} y_{k-1}^T}{y_{k-1}^T s_{k-1}} = \frac{y_{k-1} (\nabla f(x_k) - \nabla f(x_{k-1}))^T}{y_{k-1}^T (x_k - x_{k-1})}
\end{align*}
representing the information of curvature from $x_{k-1}$ to $x_k$.

While in Equ.\ref{eq:BFGS}, we use
\begin{align*}
     \frac{ \hat{y}_{k-1} \hat{y}_{k-1}^T}{ \hat{y}_{k-1}^T s_{k-1}} = \frac{ \hat{y}_{k-1} (\nabla f(\hat{x_k}) - \nabla f(\hat{x_{k-1}}))^T}{ \hat{y}_{k-1}^T (x_k - x_{k-1})}
\end{align*}
where the averaged gradients are used for calculating the curvature information. 
We will prove in lemma 1.2 that both $\nabla f(\hat{x_k})$ and $\nabla f(x_k)$ tend to be 0. 

One of the key impacting factor for our algorithm is the number of the local updates. If the number of local updates is set to 1, then $\nabla f(\hat{x_k})$ will be equal to $\nabla f(x_k)$. In this case, the proposed method is equivalent to a centralized second-order gradient descent.
When the number of local update steps is greater than 1, our method is equivalent to generating a substitute point $\hat{x}$ for $x$, then carrying out a second-order gradient descent.
In Appendix, we further verify the influence of number of local update steps in our experiments.




\textbf{Enforcing Positive-Definiteness} Finally, it is important to maintain the positive definiteness of matrix $B$ during iterations. In literature, there are mainly two solutions. The first one is through mandatory amendment. For example, in the work of \cite{2013Fast},  the matrix is forcibly guaranteed by the eigenvalue decomposition. The other solution is through the line search criteria~\cite{DBLP:journals/corr/abs-2109-02388}, which will require more communication per iteration (for checking conditions about global loss and global gradient) and will inevitably induce additional computation overhead. To ease the communication burden, we get inspiration from \cite{2016A} to design an updating process which force the curvature value and achieve better theoretical properties. Specifically, we use the following criteria by forcibly setting: 
$$\lambda < \frac{\Vert \hat{y}_{k-1}\Vert^2}{cur} < \Lambda, cur = \hat{y}_{k-1}^T s_{k-1}$$
\vspace{-4mm}



Algorithm~\ref{alg:FedSSO} demonstrates the details of our proposed FedSSO. First, the initial parameters of global model, $x_0$, are sent to each client, and clients can update the parameters locally over their private data through $\tau$ local epochs. Then the aggregated gradients $v_k$ are computed on the server based on the local updates collected from clients. 
Next we use the BFGS method to generate the approximate Hessian matrix $\hat{B}_k$ on the server and send back to the clients with the global updated model. 
Note for large-scale optimization problems, a variant of BFGS, Limited-memory BFGS (L-BFGS), can be readily adapted to further mitigate the consumption of resource~\cite{nocedal1980updating}. Fig.\ref{fig:Fed_Algorithms_Diff_on_Com} shows the comparison of the communication schemes of various first-order and second-order federated learning algorithms. It can be seen that although FedSSO is a second-order algorithm, it communicates the same information as FedAvg, whereas other second-order algorithms require multiple rounds of communication per iteration.

\begin{algorithm*}[h]
\caption{FedSSO algorithm}
\label{alg:FedSSO}
\textbf{Input}:\ number of clients $N$, $x_1$, $\hat{B}_0 = I$, $\lambda>0$, $\Lambda>0$ \\
\textbf{Output}:\ optimal weights of global model $x^*$
\begin{multicols}{2}
\begin{algorithmic}[1] 
\FOR{$k = 1 \to K$}
\STATE Server sends parameters $x_k$ to clients.
\FOR{client $i=1 \to N $ parallel }
\STATE Update $x_{k,0}^{(i)} = x_k$.
\FOR{local update $j=0 \to \tau-1$}
\STATE $x_{k,j+1}^{(i)} = x_{k,j}^{(i)} - \alpha_k \bigtriangledown f_i(x_{k,j}^{(i)}, \zeta)$
\ENDFOR

\STATE Clients send $x_{k,\tau}^{(i)}$ to server.
\ENDFOR

\STATE Aggregate $v_k = \sum_{i=1}^N p_i x_{k,\tau}^{(i)}$
\STATE Set $g(\hat{x_k}) = \frac{1}{\alpha_k \tau}(x_{k} - {v_k})$.
\STATE Set $\hat{y}_{k-1} = g(\hat{x_k})-g(\hat{x}_{k-1})$, $s_{k-1} = x_k - x_{k-1}$
\STATE Generate $\hat{B}_k$ by \textbf{BFGS Updating} 

\STATE Update $x_{k+1}
= (I-\frac{\eta_k}{\alpha_k \tau} \hat{B}_{k}^{-1})x_{k} + \frac{\eta_k}{\alpha_k \tau} \hat{B}_{k}^{-1}{v_k}$
\ENDFOR
\STATE \textbf{return} $x^* = x_K$ 
\end{algorithmic}
\columnbreak
\textbf{BFGS Updating\label{algo_hessian_update1}} process\\\\
\textbf{Input}: $\hat{y}_{k-1}$, $s_{k-1}$, $\hat{B}_{k-1}$ \textbf{  Output}: $\hat{B}_k$\\
Set $cur = \hat{y}_{k-1}^T s_{k-1}$\\
\begin{algorithmic}[1] 
\IF {$k \mod R == 0$}{
    \STATE  \textbf{return}  $I$
}\ENDIF

\IF {$\lambda < \frac{\Vert \hat{y}_{k-1}\Vert^2}{cur} < \Lambda$ is False} {
\STATE $cur = \frac{2}{\lambda + \Lambda} \Vert \hat{y}_{k-1}\Vert^2$
} \ENDIF
\STATE $\hat{B}_k=\hat{B}_{k-1}+\frac{\hat{y}_{k-1} \hat{y}_{k-1}^T}{cur}-\frac{\hat{B}_{k-1}s_{k-1}s^T_{k-1}\hat{B}_{k-1}}{s^T_{k-1}\hat{B}_{k-1}s_{k-1}}$

\textbf{return} $\hat{B}_k$

\end{algorithmic}
\end{multicols}
\end{algorithm*}
\vspace{-0.1in}

\section{Theoretical Analysis}

We list all the notations in Table.\ref{notations} on Appendix.



\textbf{Definition 5.1}[\textbf{Lighthouse Point}]
\label{lighthouse}
Here we formally define $\hat{x_k}$ point in Equ.\ref{glb_grad} as a Lighthouse point that satisfies 
\begin{center}
$
 g(\hat{x_k}) = \frac{1}{\tau} \sum_{i=1}^N p_i\sum_{j=0}^{\tau - 1} \nabla f_i(x_{k,j}^{(i)}, \zeta)
$
\end{center}
which is obtained by combining  Equ.\ref{glb_grad}, Equ.\ref{eq:Fedavg local}, .\ref{eq:Fedavg server}, and .\ref{v_k}. And its full gradient is defined as

\begin{center}\vspace{-0.05in}
$
\nabla f(\hat{x_k}) = \frac{1}{\tau} \sum_{i=1}^N p_i \sum_{j=0}^{\tau - 1} \nabla f_i(x_{k,j}^{(i)})
$
\end{center}



Next we will prove its existence in Theorem 1.1. 






\textbf{Theorem 1.1} [Existence of the Lighthouse Point]
Assume $x \in \mathbb{R}^d$, where $d$ is the dimension, and $f(x)$ and $\nabla f(x)$ are smooth and continuous.  When $N$ clients perform a FedAvg process, there exists a point $\hat{x}$ which satisfies 

\vspace{-4mm}
\begin{align*}
    \nabla f(\hat{x_k}) = \frac{1}{N \tau} \sum_{i=1}^N \sum_{j=0}^{\tau - 1} \nabla f_i(x_{k,j}^{(i)})
\end{align*}
\textbf{Remark 1} We call this point \textbf{Lighthouse} because it points out the direction of descent. Note that similar concepts are previously proposed in Scaffold~\cite{DBLP:conf/icml/KarimireddyKMRS20} and FedOPt~\cite{2020Adaptive}. In FedOPt, it is called "pseudo-gradient". 
However, we give a formal proof for its existence and further utilize it to construct a global quasi Hessian matrix and obtain a faster convergence rate. We also demonstrate its usefulness for the approximation of the global gradient from the server side without any additional cost.



Next we perform convergence analysis of our FedSSO method, for which we consider both convex and nonconvex conditions.


\textbf{Assumption 2.1} [L-Smoothness]
Each local objective function is Lipschitz smooth, that is, 
\begin{center}\vspace{-0.05in}
$f_i(y) \leq f_i(x) + \nabla f_i(x)^T(y-x) + \frac{L}{2}\Vert y -x\Vert^2, \forall i \in {1,2,...,N}$.
\end{center}

\textbf{Assumption 2.2} [Unbiased Gradient and Bounded Variance]
The stochastic gradient at each client is an unbiased estimator of the local gradient, that is $\mathbb{E}_{\zeta} [\nabla f_i(x, \zeta)] = \nabla f_i(x)$. At meantime, it has bounded variance $\mathbb{E}_{\zeta} [\Vert \nabla f_i(x, \zeta) - \nabla f_i(x) \Vert^2] \leq \sigma^2$.

\textbf{Assumption 2.3} [$\mu$-strong Convex]
Each local objective function is $\mu$-strong convex, that is 
\begin{center}\vspace{-0.05in}
$f_i(y) \geq f_i(x) + \nabla f_i(x)^T(y-x) + \frac{\mu}{2} \Vert y-x \Vert^2, \forall i \in {1,2,...,N}$.
\end{center}


\textbf{Lemma 2.3} (Enforce Positive Definiteness) Assume sequence ${\hat{B}_k}$ is generated by Equ.\ref{eq:BFGS} 
in our FedSSO algorithm. There exist constants 0 < $\underline{\kappa}$ < $\Bar{\kappa}$, such that $\{ \hat{B}_k^{-1}\}$ satisfies 
\begin{align*}
    \underline{\kappa} I \prec \hat{B}_k^{-1} \prec \Bar{\kappa} I
\end{align*}
where $I$ represent identity matrix. 

\textbf{Theorem 2.1} [Global Convergence] 
Let Assumption 2.1-2.3 hold and $\beta, \gamma, \Gamma$ be defined therein. 
Choose $\gamma^{-1} = \min \{ \frac{N L}{2 \underline{\kappa} \mu}, \frac{\mu}{2L} \}$, $\beta = \frac{2}{\mu}$, $\alpha_k = \eta_k \frac{L \Bar{\kappa}^2}{\mu \tau \underline{\kappa}} $ and $\eta_k = \frac{2}{\mu} \frac{1}{k + \gamma}$. Then, the FedSSO satisfies 
\begin{center}\vspace{-0.04in}
$\mathbb{E}_\zeta [f(x_k)] - f^* \leq \frac{\nu}{k + \gamma}$
\end{center}\vspace{-0.04in}
where $\nu = \max \{\frac{\beta^2 \Gamma}{\beta \mu - 1}, \frac{\mu}{2}(\gamma+1) \Delta_1 \}$, $\Delta_1 = \Vert x_1 - x^* \Vert^2$, and $\Gamma = \frac{L^2 \Bar{\kappa}^2 \sigma^2}{2 \mu \tau}$.

It shows that our FedSSO algorithm can reach sub-linear $\mathcal{O} (\frac{1}{k})$ convergent rate.

For nonconvex conditions, we further make the following assumption.
\\
\textbf{Assumption 3.1} [Bounded Gradient]
Follow the same assumption in FedOpt~\cite{2020Adaptive}, we assume,
\begin{equation*}
    |[\nabla f_i(x, \zeta)]| \leq G
\end{equation*}
where $G$ is a constant that bound the gradient.

\textbf{Theorem 3.1} Assume non-convex conditions 2.1,2.2, and 3.1 hold. Let $\alpha_k = \frac{1}{2\sqrt{6} \tau L k}$ and $\eta_k = \frac{1}{\sqrt{k}}$. $\underline{\kappa}$ and $\Bar{\kappa}$ are defined in lemma 2.3. $\sigma$, $G$, and $L$ are defined on assumptions. we can conclude that
\begin{align*}
    \min_{1<k<K} ||\nabla f(x_k)||^2  &\leq \mathcal{O}(\frac{f(x_1) - f(x^*)}{\sqrt{K} \underline{\kappa} (1-\varTheta)}) + \mathcal{O} (\frac{\sigma^2}{K^2 (1-\varTheta)})\\ + \mathcal{O} (\frac{\Bar{\kappa}^2 G^2 L}{K \underline{\kappa} (1-\varTheta)})
\end{align*}
where $\varTheta = 24\tau^2 \alpha^2 L^2$.

This result shows that our algorithm converges to the stationary point in the non convex case.

\section{Experiments}

\subsection{Experimental Setup}
In this section, we validate the efficiency of FedSSO via experiments from three aspects: convergent speed, communication evaluation and memory evaluation.
We compare our method with other state-of-the-art algorithms, including  first-order federated algorithms: FedSGD~\cite{2016FedSGD}, FedAvg~\cite{DBLP:conf/aistats/McMahanMRHA17}, FedAC~\cite{2020FederatedAC}, FedOpt~\cite{2020Adaptive} and Scaffold~\cite{DBLP:conf/icml/KarimireddyKMRS20}, and second-order schemes: FedDANE~\cite{DBLP:conf/acssc/LiSZSTS19} and FedNL~\cite{DBLP:journals/corr/abs-2106-02969}. 
We use grid-search method to find the optimal hyper-parameters for all algorithms. We set the parameter range for grid search as local learning rate $\alpha$ = \{0.0001, 0.0003, 0.0007, 0.001, 0.003, 0.007, 0.01, 0.03, 0.07, 0.1, 0.3, 0.7\}; global learning rate $\eta$ = \{ 0.01, 0.03, 0.07, 0.1, 0.3, 0.7, 1\}. And default values are set for other hyper-parameters in Appendix. 
We use the SGD optimizer. 
In addition, we investigate both convex models and non-convex models. For convex setting, we use a $l_2$-regularized multinomial logistic regression model (MCLR) with softmax activation and cross-entropy loss function, which has been used in~\cite{DBLP:conf/mlsys/LiSZSTS20,DBLP:conf/nips/DinhTN20}. For non-convex setting, LeNET, MLP, and CNN model are adopted. 
The experiments are conducted on several public datasets, including MNIST~\cite{mnist_dataset} and EMNIST~\cite{EMNIST_dataset}. Additional experimental results on CIFAR10~\cite{cifar10_dataset}, Shakespeare~\cite{shakespeare_dataset}, Sent140~\cite{sent140_dataset}, and LIBSVM~\cite{LIBSVM_dataset}, as well as details on data partition and the Non-IID setting are in Appendix due to space limitations. 

\subsection{Results on MNIST and EMNIST}
\label{exp:acc}
\begin{figure*}[htbp]
    \centering
    \includegraphics[width=19cm]{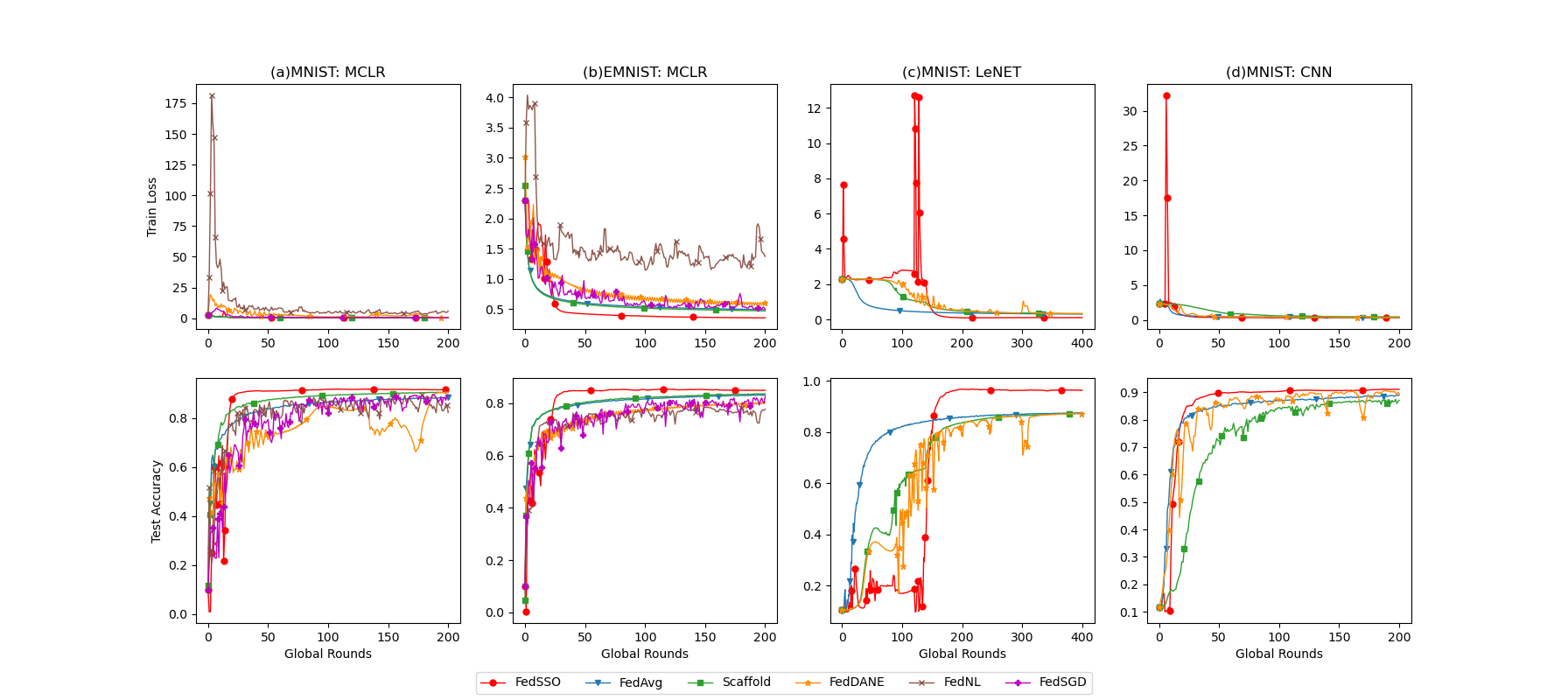}
     \vspace{-0.1in}
        \caption{The training loss and test accuracy of different federated algorithms on MNIST and EMNIST using optimal hyper-parameters.}
        \label{fig:optimal_settings}
    \vspace{-0.1in}
\end{figure*}

In order to fairly compare the performance of various algorithms, we evaluate the algorithms first using the same settings (Fig.\ref{fig:mclr_same} in Appendix) and then using their optimal settings for each algorithm Fig.\ref{fig:optimal_settings}. 


It can be seen that FedSSO is able to converge to a stationary point faster than other algorithms. It is also clear that FedSSO achieves the highest accuracy on both datasets among all algorithms with optimal hyper-parameters.
Note for non-convex models, we cannot find the proper parameters through grid-search for FedDANE and FedNL to achieve convergence, which is consistent with the conclusion of FedDANE~\cite{DBLP:conf/acssc/LiSZSTS19} and FedNL ~\cite{DBLP:journals/corr/abs-2106-02969} about non-convex cases.  

We also evaluate the convergence speed against other methods. The detailed information is listed in Table.\ref{tb:hyperparam}. 
In Table.\ref{tb:hyperparam}, we show the 
total rounds to achieve various test accuracy for various algorithms using 200 rounds of FedAvg as a reference. It can be seen that our method can achieve the corresponding accuracy with the least iteration rounds.

\begin{table*}[htb]
\caption{{Comparison of total rounds for various algorithms using their optimal hyperparameters.}}
\label{tb:hyperparam}
\begin{small}
\begin{sc}
\resizebox{\textwidth}{15mm}{
\begin{tabular}{l|l|lllll|l|l|lllll|l}
\toprule
\textbf{Algo.} & \textbf{Param} & \multicolumn{5}{c|}{\textbf{Test Acc on Mnist for MCLR}}&\textbf{Total bits} &\textbf{Param} & \multicolumn{5}{c}{\textbf{Test Acc on EMnist for MCLR}} &\textbf{Total bits} \\
Round & ~& 0.40 & 0.60& 0.80 & 0.88& 0.90 & \qquad  (B) & ~& 0.40 & 0.60& 0.80 & 0.83& 0.84 & \qquad (B) \\
\midrule
FedAvg & $\alpha$=0.03 &  1  &  2  & 27 & \textbf{200}  & - &12560000 & $\alpha$=0.03  & 0  &  2  & 50 & \textbf{200} & - &12560000 \\
Scaffold & $\alpha$=0.03  & 1  &  5  & 14 & 71 & 149 &\ \ 8917600 & $\alpha$=0.03  & 1  &  2  & 46 & 147 & - & 18463200\\
FedDANE&  $\alpha$=0.003 & 2  &  30  & 180 & 186 & 189 &22361600 & $\alpha$=0.001 & 0  &  8  & 200 & - & - & 25120000\\
FedNL & $\alpha$=0.001 & 5  &  13  & 143 & 181 & - &11366800 & $\alpha$=0.7  & 5  &  9  & - & - & -  & - \\
FedSSO & $\alpha$=0.001,$\eta$=1 &  14  &  14  & 17 & \textbf{20} & 24 &\ \ 1256000 &  $\alpha$=0.003,$\eta$=1 & 1 &  17  & 24 & \textbf{26} & 30 & \ \ 1632800 \\
\midrule
\textbf{Algo.} & \textbf{Param} & \multicolumn{5}{c|}{\textbf{Test Acc on Mnist for CNN}} & \textbf{Total bits} &\textbf{Param} & \multicolumn{5}{c}{\textbf{Test Acc on EMnist for CNN}} & \textbf{Total bits}\\
Round & ~& 0.40 & 0.60& 0.80 & 0.88& 0.91 & \qquad (B) & ~& 0.40 & 0.60& 0.80 & 0.81& 0.83 & \qquad (B) \\
\midrule
FedAvg & $\alpha$=0.3 &  6  &  9  & 20 & \textbf{200} & - &\ \ 9968000  & $\alpha$=0.1  & 5  &  12  & 196 & \textbf{200} & - & \ \ 9968000 \\
Scaffold & $\alpha$=0.01  & 24  &  33  & 82 & - & - & \ \ 8173760 & $\alpha$=0.07  & 6  &  12  & 196 & 198 & - & 19736640 \\
FedSSO & $\alpha$=0.07,$\eta$=0.3 &  10  &  13  & 34 & \textbf{40} & 192 &\ \ 1993600 &  $\alpha$=0.003,$\eta$=1 & 14 &  17  & 46 & \textbf{67} & 191 & \ \ 3339280 \\
\bottomrule
\end{tabular}}
\end{sc}
\begin{tablenotes}
{\footnotesize \item  '-' indicates that this algorithm cannot reach this accuracy in 200 rounds of training process. } 
\end{tablenotes}
\end{small}
\vskip 0.03in
\end{table*}


\subsection{Communication evaluation}
The communication cost of FL algorithms depends on both the communication rounds and total bits communicated. 
In Table.\ref{tb:com} we provide theoretical analysis on the total bits per communication round for each algorithm, where we use FedAvg as baseline and denote its bits per round as $n_c$.


\begin{table}[htb]
\centering
\caption{Comparison of communication cost and memory usage.}
\label{tb:com}
\vspace{-0.1in}
\vskip 0.015in
\begin{small}
\begin{sc}
\begin{tabular}{l|c|c|c}
\toprule
\textbf{Algo} & \textbf{Com P.R} & \textbf{Server Mem} & \textbf{Client Mem} \\
\midrule
FedAvg & {$n_{c}$} & {$n_{m}$} & {$n_{m}$} \\
Scaffold &$2n_{c}$&$2n_{m}$&$2n_{m}$ \\
FedDANE& $2n_{c}$&$2n_{m}$&$2n_{m}$  \\
FedNL & $n_{c}^2$&$2n_{m}^2+2n_{m}$&$n_{m}^2+2n_{m}$    \\
FedSSO & {$n_{c}$} & $n_{m}^2+4n_{m}$ & {$n_{m}$}   \\
\bottomrule
\end{tabular}
\end{sc}
\begin{tablenotes}
{\footnotesize \item 'COM P.R' represents communications per round. }
{\footnotesize \item '$n_c$' is denoted as its bits per round by FedAvg and as a baseline. }
{\footnotesize \item '$n_m$' is denoted as its required memory on server by FedAvg and as a baseline.}
\end{tablenotes} 
\end{small}
\end{table}

From Table.\ref{tb:com}, we see that FedSSO only communicates the same level of bits as FedAvg. However, in FedDANE, an additional communication round is added for transmitting global gradients; In Scaffold, clients need to send both local models and correction terms to server. As a result, the total bits are doubled in FedDANE and Scaffold. In FedNL, the compressed updates of Hessian matrices need to be uploaded to the server in addition to the gradients. Here we report the theoretical estimation of all algorithms without any compression. 
We also report the total bits (B as unit) for the optimal setting in Table.\ref{tb:hyperparam} TOTAL BITS column.
Its calculation formula follows that: 

Total bits = Communication per round * rounds.

In practice, compression techniques can be applied to these algorithms to further reduce communication cost, but additional computing cost may occur.

\subsection{Memory evaluation}
The memory usage of both server and clients by these algorithms are also estimated in Table.\ref{tb:com}. As a baseline, the memory required on server in FedAvg is denoted as $n_{m}$. Additional updates of correction terms are introduced into Scaffold, which doubles the required memory in computation on both clients and server. In FedDANE, aggregation operations of gradients and models are both conducted at server, and clients are required to store global models, global gradients, and local gradients simultaneously to obtain optimal local models. In FedSSO, server updates the global model using the approximated global gradient, and it needs to store the global model and estimated gradients in both previous and current round. FedNL has the highest memory requirements because the computation of exact Hessian matrices in each client is required and the learned Hessian matrix in server needs to be updated using the aggregated first-order gradients. 
In summary, our proposed FedSSO has the lowest memory requirements on the client sides, and the main resource consumption is on the server side.




\section{Conclusion}

We present FedSSO, a server-side second-order optimization algorithm for FL which adopts a server-side estimation of global gradients and Hessian matrix to update the global model. We provide theoretical guarantee for the convergence of FedSSO, and analyze the convergence rate. Extensive experimental evaluations against state-of-the-art FL algorithms show that our proposed FedSSO can outperform other counterparts in both convex and non-convex settings. Our method enjoys fast convergence of second-order algorithms while requiring no additional computation or communication burden from clients as compared to FedAvg, which makes it practical for FL implementations. We hope our work can shed light on future work for the server-side second-order optimization algorithms. 



%


\bibliography{aaai23}



\newpage
\onecolumn

\section{Appendices}

The appendices are structured as follows.

\section{Additional Experiments and setup details}

\subsection{Additional setup details}
\subsubsection{datasets on Non-IID setting}
\label{appendix_experiment_data}
In our experiments, the setting of Non-IID data is automatically generated through the open source framework PFL-Non-IID (\url{https://github.com/TsingZ0/PFL-Non-IID}).
For the Non-IID setting, we mainly consider three aspects: the amount of data on each client, the distribution of data categories on each client, and the data categories on each client. For LIBSVM dataset, we use its ijcnn data, and the details of data distribution are shown in Table.\ref{tb:client_distribution}:

\begin{table*}[htb]
\caption{LIBSVM \# Label and size of samples in clients.}
\label{tb:client_distribution}
\vspace{0.01in}
\begin{center}
\begin{small}
\begin{tabular}{l|l}
\toprule
\textbf{No.client} & \textbf{label and size of samples on client [label, size]}\\
\midrule
client 1 & [1,1298],[5,6313],[8,4038],[9,3396] \\
client 2 & [0,580],[7,7293],[8,2787] \\
client 3 & [0,6323],[1,6579],[2,6990],[3,7141],[4,6824],[6,6876],[9,3562] \\
\bottomrule
\end{tabular}
\end{small}
\end{center}
\vskip 0.01in
\end{table*}



The setup for other datasets are similar to LIBSVM, except that the number of clients for MNIST and EMNIST datasets is set to 20 and the number of clients for Cifar10 dataset is 10. 
For Shakespeare dataset, we partition the dataset into 10 clients without shuffling to mimic a non-iid setting. The task is to predict the next character and the length of each segment is fixed to 50. 
For Sent140 dataset (\url{http://cs.stanford.edu/people/alecmgo/trainingandtestdata.zip}), Glove (\url{http://nlp.stanford.edu/data/wordvecs/glove.twitter.27B.zip}) is used to extract embeddings. The length of segment is set to 200 and data is partitioned into 10 clients without shuffling as non-iid setting. 

For all of datasets, the train data and test data is split with a ratio 0.75 on each client side. For all algorithms and all settings, we mainly evaluate the loss and accuracy for the global model. 

Our data and experimental results are available on GitHub( 
 \url{https://github.com/baobaoyeye/FedSSO-Datasets} ). All the methods are implemented in Torch 1.8.2+cu111 running on NVIDA 40GB GPU memory.
 
\subsubsection{Hyper-parameters setting}
For all algorithms, we use the gird search to tune its local learning rate $\alpha$ and global learning rate $\eta$. 
For our FedSSO, we set $R = 200$, $\lambda = 0.0001$, and $\Lambda = 9999$, that are used to enforce positive definiteness. the $\lambda$ and $\Lambda$ are important to handle the non convex tasks, especially. Empirically, when the $\Lambda$ is set to big, it may lead to unstable.
For FedOpt, we follow its work~\cite{2020Adaptive} to set its momentum parameters with 0.9 and 0.001 as default values.
For FedDane, $\mu$ is set to 0.001 as default value.

\subsubsection{Implementations}
According to our BFGS updating process, the $B$ is directly generated. However, we need its inverse formal. For handling this issue, we provide two versions for solving its inverse. For the one version, we use the solver in torch to solve a equation, which can directly get $\hat{B}_k^{-1} \nabla f(\hat{x_k})$ solution. For another version, we use the inverse formal of DFP, that is, the dual formal of BFGS equation, which can directly get its inverse. 
Empirically, for small tasks, the first version is enough to handle. And for big tasks, the second version is suggested to adopt, considering efficiency.

\subsection{More Results on MNIST and EMNIST}
\begin{figure*}[htb]
    \center{\includegraphics[width=19cm]  {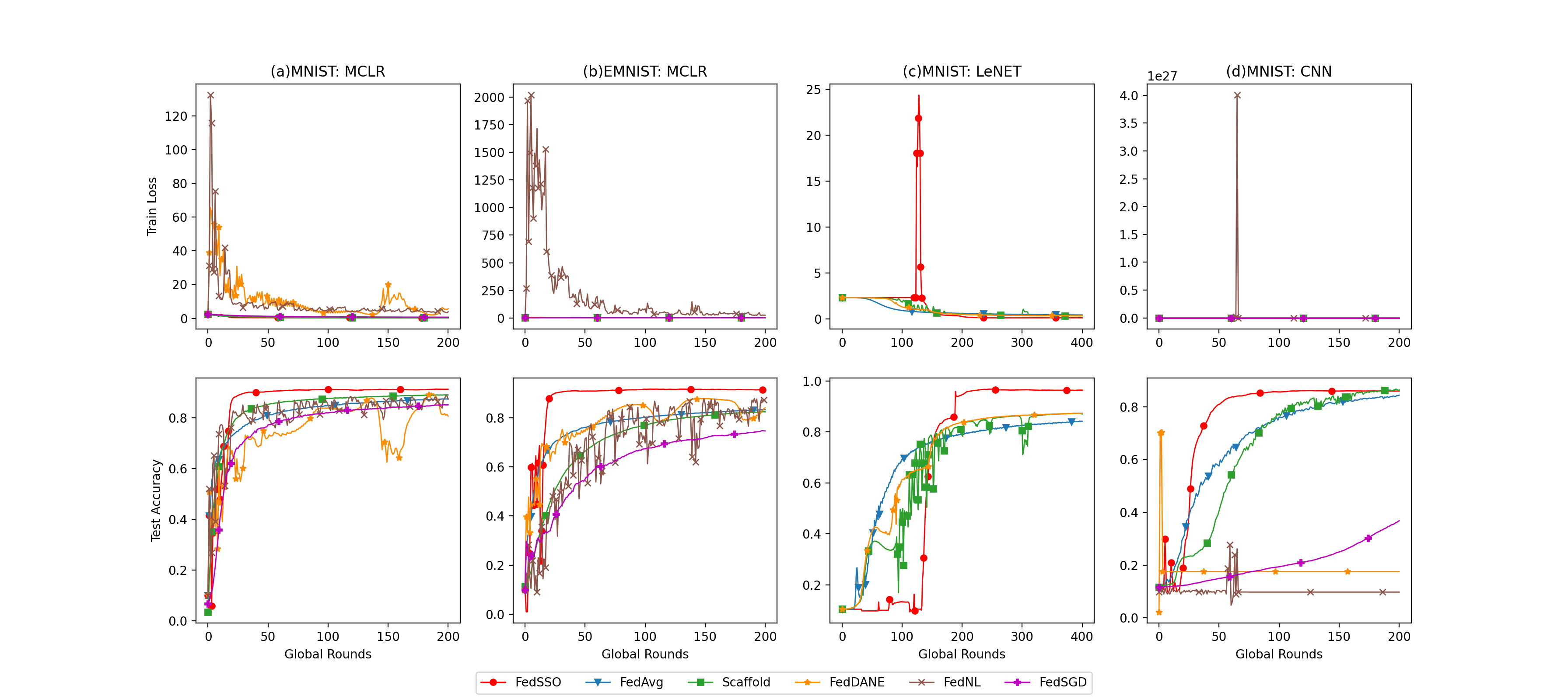}} 
        \caption{The training loss and test accuracy of different federated algorithms using the same hyper-parameters setting($BS=100$, $\tau=5$ and $\alpha = 0.001$ for convex model, $BS=100$, $\tau=5$, $\alpha = 0.01$ for non-convex model). } 
        \label{fig:mclr_same}
        \vskip -0.03in
        \vspace{-0.1in}
\end{figure*}

\ \
Fig.\ref{fig:mclr_same} shows the accuracy and training loss of different algorithms 
using the same hyper-parameters ($BS=100$, $\tau=5$ and $\alpha = 0.001$ for convex model, $BS=100$, $\tau=5$, $\alpha = 0.01$ for non-convex model). 

We can see that for convex models, FedSSO is less stable at first due to the large gap between initial model and optimal model, but it reaches to a lower level of training loss and higher accuracy compared to other algorithms eventually. For non-convex models, all algorithms need more rounds before obtaining the optimal and stationary model, and the period of instability of FedSSO becomes longer. From Fig.\ref{fig:mclr_same} it is clear that FedSSO can outperform other first-order and second-order algorithms.

\subsection{Comparison with Other Algorithms on Cifar10, Shakespeare and Sent140}
\begin{figure*}[htbp]
    \centering
    \includegraphics[width=19cm]{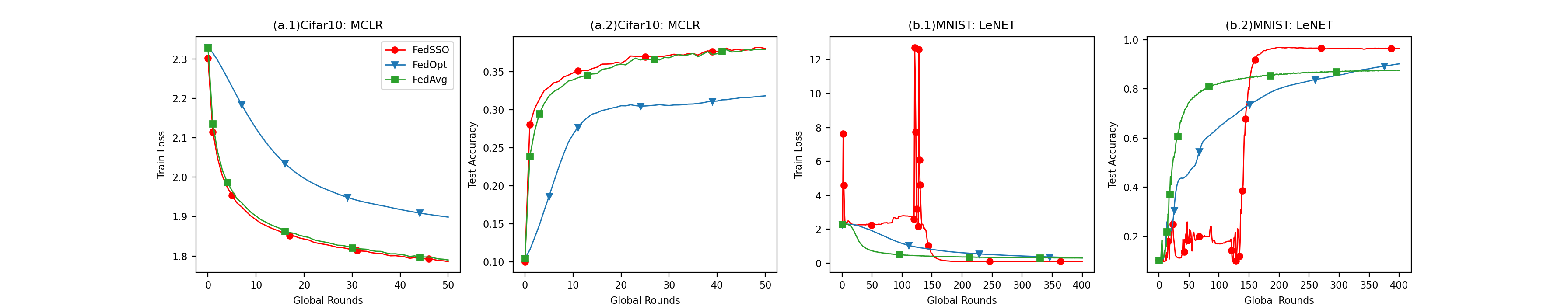}
    
        \caption{The training loss and test accuracy of FedSSO,FedOpt and FedAvg using their optimal hyper-parameters setting.}
        \label{fig:optimal_cifar10}
\end{figure*}

\begin{figure*}[htb]
     \begin{subfigure}[b]{0.24\linewidth}
         \centering
         \includegraphics[width=1.8in]{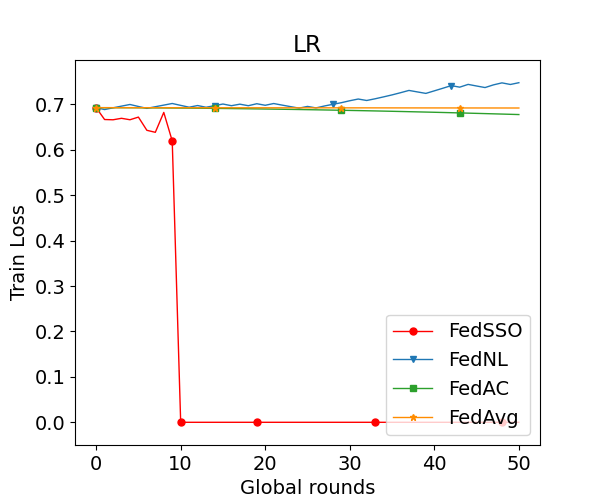}
         \caption{clients: 3 \# $\alpha$: 0.001.}
         \label{fig:mclr_same_a}
     \end{subfigure}
     \begin{subfigure}[b]{0.24\linewidth}
         \centering
         \includegraphics[width=1.8in]{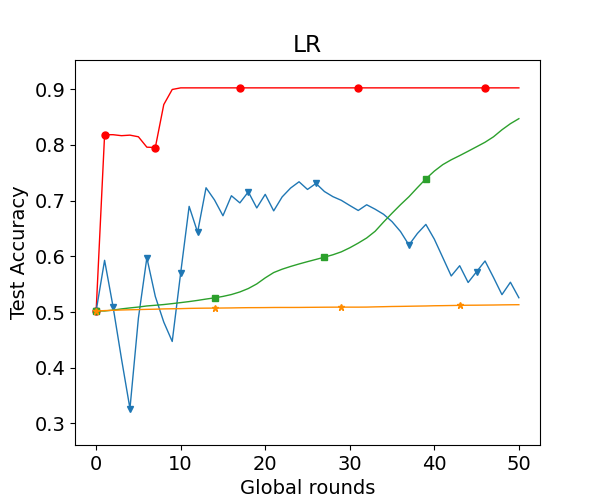}
         \caption{clients: 3 \# $\alpha$: 0.001.}
         \label{fig:mclr_same_20a}
     \end{subfigure}
     \begin{subfigure}[b]{0.24\linewidth}
         \centering
         \includegraphics[width=1.8in]{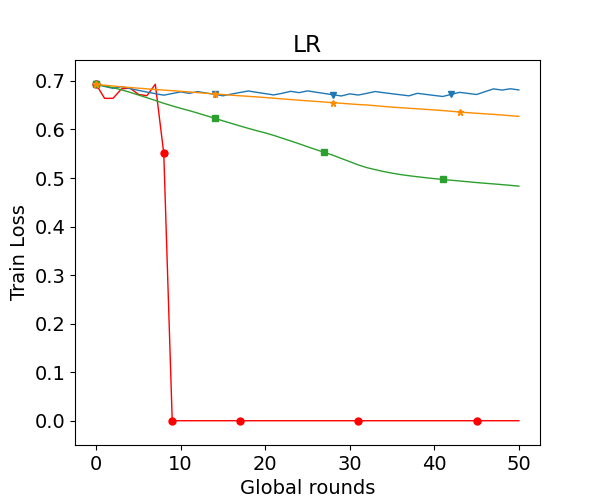}
         \caption{clients: 3 \# optimal.}
         \label{fig:cnn_a}
     \end{subfigure}
     \begin{subfigure}[b]{0.24\linewidth}
         \centering
         \includegraphics[width=1.8in]{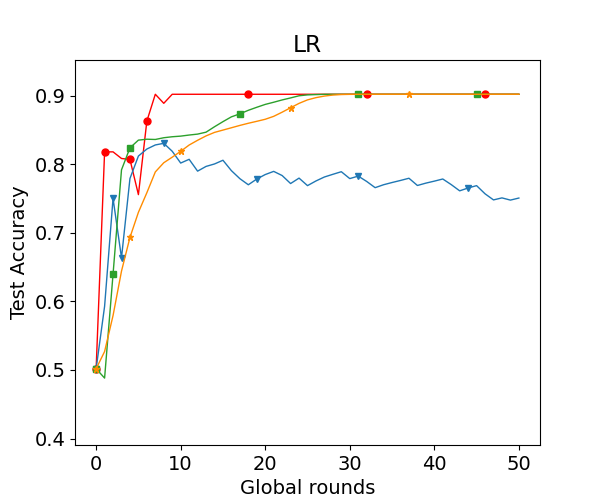}
         \caption{clients: 3 \# optimal.}
         \label{fig:cnn2_a}
     \end{subfigure}
    \caption{The training loss and test accuracy of FedSSO, FedNL, FedAC and FedAvg on LIBSVM for LR model using the same parameter settings (a,b) and optimal hyper-parameter settings (c,d).} 
    \label{fig:optimal_libsvm}
    \vskip -0.03in
    \vspace{-0.1in}
\end{figure*}

\begin{figure*}[htb]
     \begin{subfigure}[b]{0.24 \linewidth}
         \centering
         \includegraphics[width=1.8in]{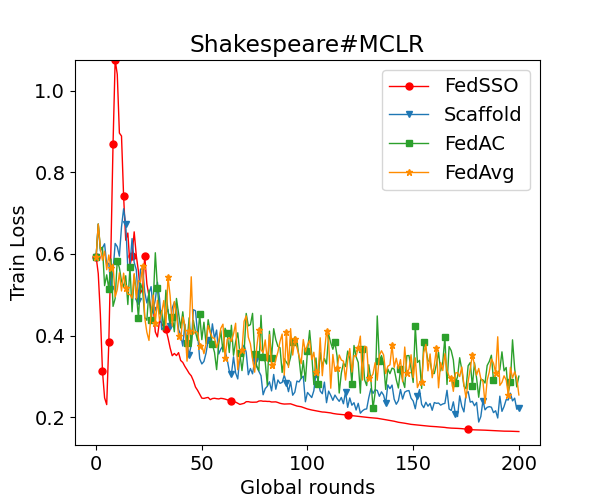}
         \caption{clients: 10 \# $\alpha$: 0.003.}
         \label{fig:mclr_same_a}
     \end{subfigure}
     \begin{subfigure}[b]{0.24 \linewidth}
         \centering
         \includegraphics[width=1.8in]{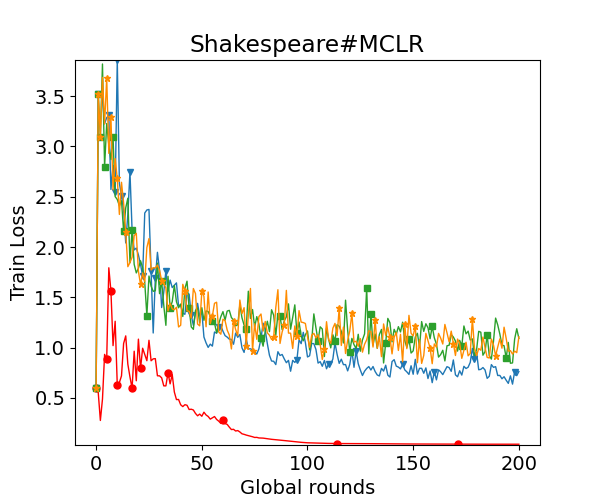}
         \caption{clients: 10 \# optimal.}
         \label{fig:mclr_same_20a}
     \end{subfigure}
     \begin{subfigure}[b]{0.24 \linewidth}
         \centering
         \includegraphics[width=1.8in]{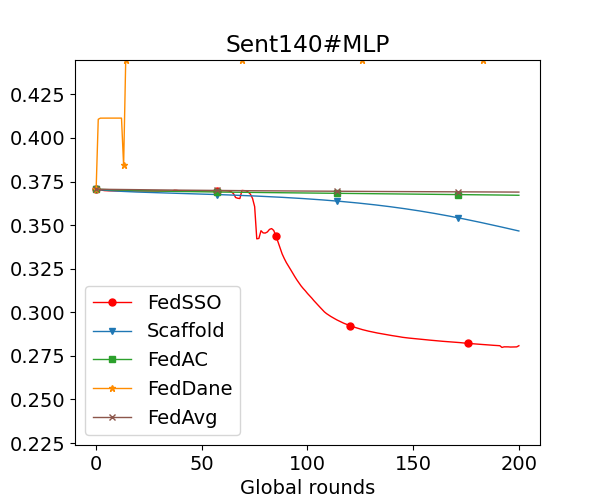}
         \caption{clients: 10 \# $\alpha$: 0.03.}
         \label{fig:cnn_a}
     \end{subfigure}
     \begin{subfigure}[b]{0.24 \linewidth}
         \centering
         \includegraphics[width=1.8in]{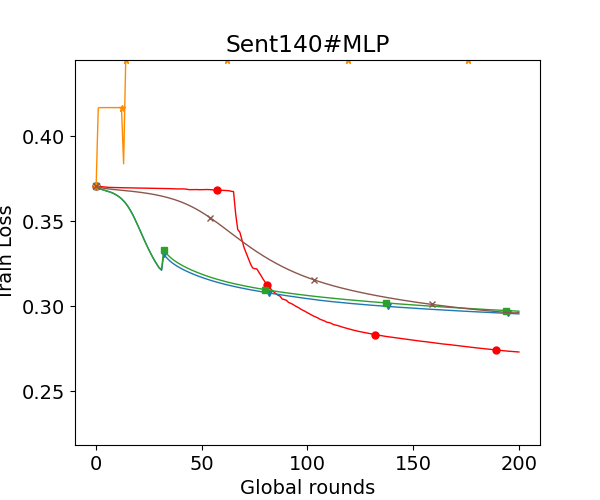}
         \caption{clients: 10 \# optimal.}
         \label{fig:cnn2_a}
     \end{subfigure}
     \begin{subfigure}[b]{0.24 \linewidth}
         \centering
         \includegraphics[width=1.8in]{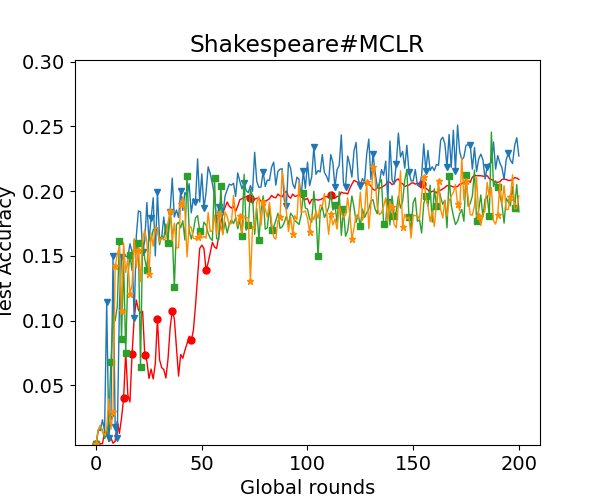}
         \caption{clients: 10 \# $\alpha$: 0.003.}
         \label{fig:mclr_same_b}
     \end{subfigure}
     \begin{subfigure}[b]{0.24 \linewidth}
         \centering
         \includegraphics[width=1.8in]{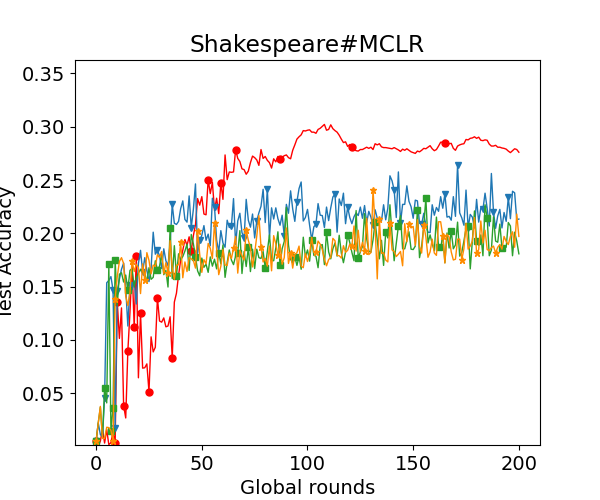}
         \caption{clients: 10 \# optimal.}
         \label{fig:mclr_same_20b}
     \end{subfigure}
     \begin{subfigure}[b]{0.24 \linewidth}
         \centering
         \includegraphics[width=1.8in]{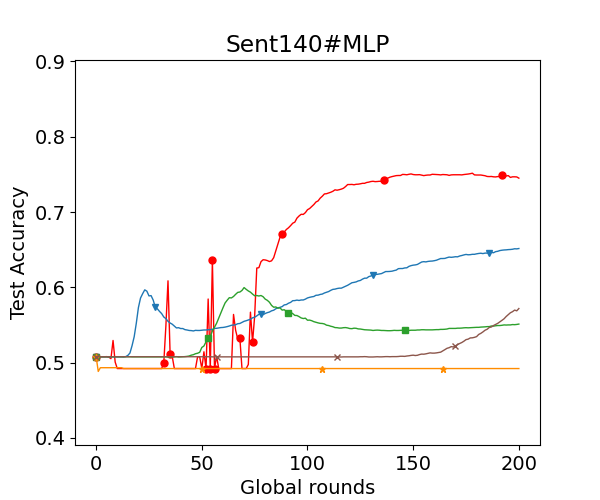}
         \caption{clients: 10 \# $\alpha$: 0.03.}
         \label{fig:cnn_b}
     \end{subfigure}
     \begin{subfigure}[b]{0.24 \linewidth}
         \centering
         \includegraphics[width=1.8in]{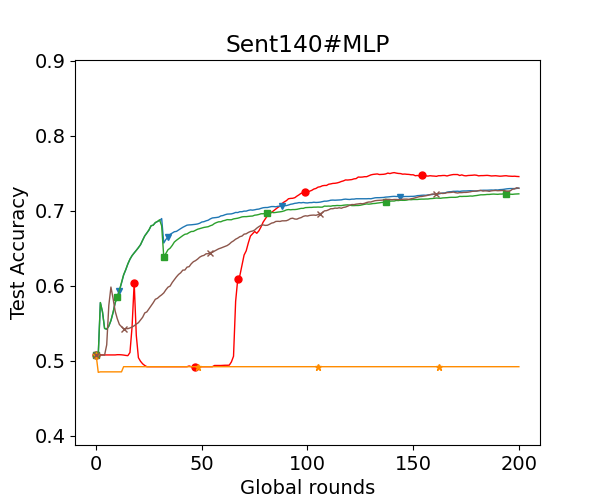}
         \caption{clients: 10 \# optimal.}
         \label{fig:cnn_b}
     \end{subfigure}
    \caption{The training loss (upper) and test accuracy (lower) for different algorithms on Shakespeare and Sent140 datasets with the same (a and e, c and g) and optimal hyperparameters setting(b and f, d and h). } 
    \label{fig:nlp}
\end{figure*}

FedOpt~\cite{2020Adaptive} is a novel first-order algorithm which use the similar definition of average gradient with ours. FedAC~\cite{2020FederatedAC} is an accelerate techniques used in FL.
In this section, we show more evaluations against these algorithms on additional datasets. 
For FedOpt, we use its FedAdaGrad implementation. 
The results are shown in Fig.\ref{fig:optimal_cifar10}, Fig.\ref{fig:optimal_libsvm},and Fig.\ref{fig:nlp}.


It can be seen from Fig.\ref{fig:optimal_cifar10} that our method can still achieve faster convergence than FedOpt, whether on Cifar10 or using LeNET model~\cite{lenet_cite}.


Fig.\ref{fig:optimal_libsvm} shows the results on a binary classification task on LIBSVM dataset by comparing various Quasi-Newton methods including FedAC. 

    

It can be seen from Fig~\ref{fig:optimal_libsvm} that FedAC can also achieve good convergence, and FedSSO still achieves faster convergence than other methods. This is consistent with the conclusion that the second-order method is generally faster than the first-order method. 




We also conduct experiments on the larger Shakespeare and Sent140 datasets, for which MCLR and MLP~\cite{mlp_cite} models are adopted respectively. 
The results are shown in Fig.\ref{fig:nlp}. 


Fig~\ref{fig:nlp} also shows faster convergence for our FedSSO. For non convex models, our method generally need more exploration at the initial stage. For convex model, the exploration of our method in the initial stage is not so obvious. The reason may be the different space of the loss function.
Another noteworthy phenomenon is that it is usually difficult to train second-order FedDane and FedNL. 
For MLP model trained on FedDane, it's difficult to be convergent, like Fig~\ref{fig:mclr_same}. 
Our experiment results are consistent with the work~\cite{DBLP:conf/acssc/LiSZSTS19} and ~\cite{DBLP:journals/corr/abs-2106-02969}, which hold this view.
Because in the process of training, its loss value may increase infinitely to become 'Nan' value. In fact, this may be a common phenomenon for the second-order methods. For our method, we usually limit the Positive Definiteness parameter setting ($\lambda$, $\Lambda$) to avoid this problem, especially on non convex tasks.

\subsection{Impact of Number of Local updates}
\begin{figure}[h]
     \centering
    
     \begin{subfigure}[b]{0.4\textwidth}
         \centering
         \includegraphics[width=\textwidth]{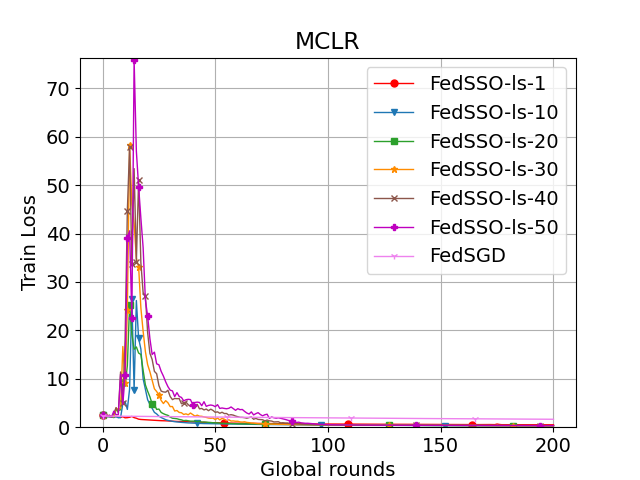}
         \caption{Train Loss on MNIST \\ for different local step.}
         \label{fig:mclr_opt_20a}
     \end{subfigure}
     \begin{subfigure}[b]{0.4\textwidth}
         \centering
         \includegraphics[width=\textwidth]{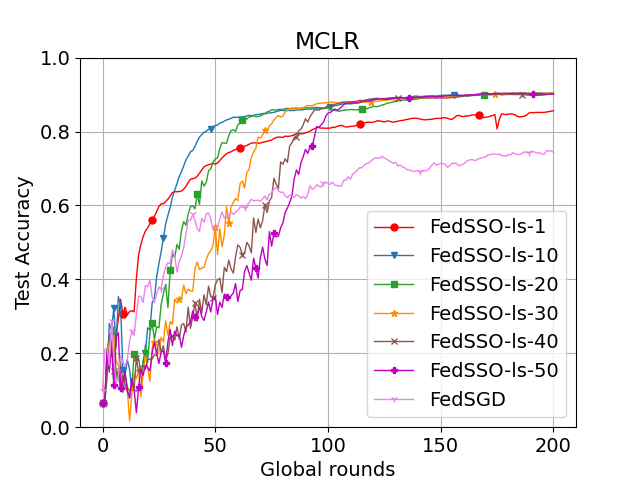}
         \caption{Test Accuracy on MNIST \\ for different local step.}
         \label{fig:mclr_opt_20a_f}
     \end{subfigure}
     
     \caption{The different local update step and 20 clients for our FedSSO algorithm on MNIST with same setting }
     \label{diffrent_local_step}
\end{figure}

At each iteration, the descent process of our FedSSO algorithm is divided into two steps: a Federated process and a Quasi-Newton process. When the number of local updates is 1, our method will degenerate into a second-order stochastic Quasi-Newton algorithm. Fig.\ref{diffrent_local_step} evaluates the impact of the number of local updates. As the the number of local updates increases, we observe that the convergence becomes faster at first, but eventually becomes prolonged when the number of local updates is too large. This shows that the effectiveness of the Lighthouse point and our methods depends on choosing a reasonable range for the number of local updates, which will result in a better descent direction for the quasi Newton process. When the number of local updates is small, the acceleration of the federated process is not obvious. When the number of local updates is too large, the resulting lighthouse is too far from optimal, which will have adverse effect on the convergence.  

\section{Preliminary}
\label{preliminary_appendix}

\subsection{Notations}
We list the notations used in Table~\ref{notations} below:

\begin{table}[h] 
\caption{Notations}  
\centering
\begin{tabular}{ll}   
\toprule
Notations   &    Definition    \\
\midrule  
N           & number of clients \\
$\tau$      & local update steps \\
k           & round number \\
$\eta$, $\alpha$    & global step length, local step length\\
$B$             & approximate Hessian matrix by BFGS \\
$\hat{B}$       & approximate BFGS Hessian matrix \\
$R$             & constant used in BFGS update process \\
$\Bar{\kappa}, \underline{\kappa}$   & upper and lower bound for $\hat{B}$ \\
$L, \mu$        & L-smooth, $\mu$-convex \\
$I$             & Identity matrix \\
$y_k, s_k$      & used in BFGS  \\
$\lambda, \Lambda$     & bound for curvature \\
$p_i$           & model weight, that is $\sum_i^N p_i = 1$ \\
$\sigma$        & assumption bound for variance \\
$G$             & assumption bound for gradient \\
$x$, $\hat{x_k}$         & model parameter, Lighthouse point        \\ 
$x^\mathrm{T}$  &  Transpose x \\
$v_k$               &  aggregation models after k-times\\
$x_{k,j}^{(i)}$     &  $i$-th client's updates on  step $j$ on round $k$ \\
$f(x)$, $f_i(x)$        & total and i-th client objective function  \\
$\zeta$                 & stochastic variable \\
$\nabla f_i (x_{k,j}^{(i)}, \zeta)$         & $i$-th client's stochastic gradient on $x_{k,j}^{(i)}$      \\
$\nabla f(x)$       & gradient \\
$\nabla f(x, \zeta),g(x)$  &   Both are stochastic gradient \\
$f'(x)$    & one-dimension gradient \\
$BS$       & experiment setting: the batch size \\

\bottomrule
\end{tabular}
\label{notations}
\end{table}

Then, we state the theorems, formulas and knowledge used for subsequent proof below.

\subsection{Function properties}
If the function is $\mu$-strong convex, we always obtain that:
\begin{equation}
\label{strong_convex1}
    f(y) \geq f(x) + \nabla f(x)^\mathrm{T} (y - x) + \frac{1}{2} \mu \Vert y - x \Vert^2,
\end{equation}
\begin{equation}
    \Vert \nabla f(y) - \nabla f(x) \Vert \geq \mu \Vert y - x \Vert
\end{equation}

If the function satisfy the Lipschitz condition, we always obtain that:
\begin{equation}
\label{lipschitz_condition1}
    f(y) \leq f(x) + \nabla f(x)^\mathrm{T} (y - x) + \frac{1}{2} L || y - x ||^2,
\end{equation}
\begin{equation}
\label{lipschitz_gradient}
    || \nabla f(y) - \nabla f(x) || \leq L || y - x ||
\end{equation}

\subsection{Inequality Properties}
According to Cauchy inequality, we can get that:
\begin{equation}
    \label{cauchy_inequality}
    \Vert \sum_{i=1}^{n} x_i \Vert^2 \leq n \sum_{i=1}^{n} \Vert x_i \Vert^2,
\end{equation}

\begin{equation}
\label{cauchy_multiple}
    x^\mathrm{T} y \leq \Vert x \Vert \Vert y \Vert
\end{equation}




\subsection{Algorithm Details}
\label{appendix_algo_details}
We describe the common FedAvg algorithm here.

\begin{algorithm}[h]
\caption{FedAvg algorithm}
\label{alg:FedAvg}
\textbf{Input}:~number of clients: $N$, \\ 
number of samples on client $i$:~$n_i$, \\
initial model: $x_0$\\
\textbf{Output}:Optimal global model $x^*$

\begin{algorithmic}[1] 
\FOR{$k = 0 \to K-1$}
\STATE Server sends parameters $x_k$ to clients.
\FOR{client $i=1 \to N$}
\STATE Update $x_{k,0}^{(i)} = x_k$.
\FOR{local update $j=0 \to \tau-1$}
\STATE $x_{k,j+1}^{(i)} = x_{k,j}^{(i)} - \alpha \nabla f_i(x_{k,j}^{(i)}, \zeta)$
\ENDFOR
\STATE Clients send $x_{k,\tau}^{(i)}$ to server.
\ENDFOR
\STATE Server aggregate $x_{k+1} = \frac{1}{\sum_{j=1}^N n_j}\sum_{i=1}^N n_i x_{t,\tau}^{(i)}$
\ENDFOR
\STATE \textbf{return} $x^* = x_{K}$
\end{algorithmic}
\end{algorithm}

According to the FedAvg algorithm \ref{alg:FedAvg}, we describe its one-round descent process as follow:

ith-Client: Federated process

\hspace{0.5cm} 1) $x_{k,0}^{(i)} = x_k$, where $x_k$ is received from server.

\hspace{0.5cm} 2) local upstate $\tau$ steps:  $x_{k,\tau}^{(i)} = x_{k,0}^{(i)} - \alpha_k \sum_{j=0}^{\tau-1} \nabla f_i(x_{k,j}^{(i)}, \zeta)$.

\hspace{0.5cm} 3) send $x_{k,\tau}^{(i)}$ back to server.

Server: Federated process

\hspace{0.5cm} 1) aggregate $x_{k+1} = \sum_{i=1}^N p_i x_{k,\tau}^{(i)}$.

\hspace{0.5cm} 2) send $x_{k+1}$ to clients.

In order to make the following theoretical analysis more intuitive, we describe our FedSSO algorithm \ref{alg:FedSSO} on one-round descent as federated and Quasi-Newton processes:

ith-Client: Federated process

\hspace{0.5cm} 1) $x_{k,0}^{(i)} = x_k$, where $x_k$ is received from server.

\hspace{0.5cm} 2) local upstate $\tau$ steps:  $x_{k,\tau}^{(i)} = x_{k,0}^{(i)} - \alpha_k \sum_{j=0}^{\tau-1} \nabla f_i(x_{k,j}^{(i)}, \zeta)$.

\hspace{0.5cm} 3) send $x_{k,\tau}^{(i)}$ back to server.

Server: Quasi-Newton process

\hspace{0.5cm} 1) aggregate $v_k = \sum_{i=1}^N p_i x_{k,\tau}^{(i)}$.

\hspace{0.5cm} 2) generate $\hat{B}_k$ By \textbf{Hessian update process}.

\hspace{0.5cm} 3) update $x_{k+1}
= (I-\frac{\eta_k}{\alpha_k \tau} \hat{B}_{k}^{-1})x_{k} + \frac{\eta_k}{\alpha_k \tau} \hat{B}_{k}^{-1}{v_k}$, which is also equal to $x_{k+1} = x_k - \eta_k \hat{B}_k^{-1} g(\hat{x_k})$.

\section{Lighthouse Analysis}
\label{lighthouse_appendix}


We use \textbf{Lighthouse} to represent $\hat{x_k}$, that is the point where the average gradient is located. We give proof for Lighthouse point existance.

\textbf{Difficulty Analysis: }
The difficulty of proof is that this is a federal training process, that is, multiple clients first locally update many steps by local biased gradient, and then aggregate.
For the mean value theorem, if it is only a centralized training process, this conclusion will be obvious. However, for multiple clients to locally update many steps with biased gradient and then aggregate, there will be great obstacles.

The proof idea is to use the mean value theorem twice.
First, there will be a point for the local update process of each client.
Then, from the perspective of aggregation, for the point where each client exists, we can use the mean value theorem again to get a global point.

We first give this proof from the perspective of one dimension through Theorem 1.1, which requires Lemma 1.1.
Then we extend the conclusion to multi-dimension, in Corollary 1.1.

\textbf{Lemma 1.1}
Assume $x \in \mathbb{R}$, $f: \mathbb{R} \to \mathbb{R}$, and $ f'(x)$ is smooth and continuous in $[x_k, x_{k+\tau}]$. Then there always exist a $\hat{x_k} \in [x_k, x_{k+\tau}]$, makes below hold,
$$
 f'(\hat{x_k}) = \frac{1}{\tau} \sum_{i=0}^{\tau}  f'(x_{k+i})
$$

proof.
Due to $x$ is one-dimension, we use $f'(x)$ to represent $\frac{\partial f}{\partial x}$ for convenience.

Construct a auxiliary function 
$$
h(x) = \frac{1}{\tau} \sum_{i=0}^{\tau}  f'(x) - \frac{1}{\tau} \sum_{i=0}^{\tau}  f'(x_{k+i}) = \frac{1}{\tau} \sum_{i=0}^{\tau} ( f'(x) - \frac{1}{\tau} \sum_{i=0}^{\tau}  f'(x_{k+i}))
$$
Consider that, there always exist a $x_1 \in [x_k, x_{k+\tau}]$, and $x_2 \in [x_k, x_{k+\tau}]$, makes below hold,
$$
h(x_1) < 0, 0 < h(x_2)
$$
Then, according to the intermediate value theorem, there must exist a $\hat{x_k} \in [x_k, x_{k+\tau}]$, makes below hold
$h(\hat{x_k}) = 0$, which proves 
$$
 f'(\hat{x_k}) = \frac{1}{\tau} \sum_{i=0}^{\tau}  f'(x_{k+i})
$$

\textbf{Theorem 1.1}[Lighthouse]
\label{lemma_lighthouse_new_2}
 Consider stimulate federated algorithm Fedavg, but update by true gradient. Assume $x \in \mathbb{R}$, $f(x) = \frac{1}{n} \sum_{i=1}^{n} f_i (x)$, $f'(x) = \frac{1}{n} \sum_{i=1}^{n} f_i '(x)$, and $f_i ' (x)$ is smooth and continuous.
When local update $\tau$ steps, makes below hold
$$
 f'(\hat{x_k}) = \frac{1}{n \tau} \sum_{i=1}^{n} \sum_{j=0}^{\tau} f_i' (x_{k,j}^{(i)})
$$

proof.
Define 
$$
 f' = \frac{1}{n \tau} \sum_{i=1}^{n} \sum_{j=0}^{\tau}  f_i' (x_{k,j}^{(i)}) = \frac{1}{n} \sum_{i=1}^{n} \frac{1}{\tau} \sum_{j=0}^{\tau}  f_i '(x_{k,j}^{(i)})
$$
According to lemma 1.1, we know there always exist a $\hat{x_k}^{(i)}$, makes below hold
$$
 f_i '(\hat{x_k}^{(i)}) = \frac{1}{\tau} \sum_{j=0}^{\tau}  f_i' (x_{k,j}^{(i)})
$$
Then, substitute it into above formula, we know that
$$
 f' = \frac{1}{n} \sum_{i=1}^{n}  f_i' (\hat{x_k}^{(i)})
$$
Again, consider intermediate value theorem, and construct auxiliary function
$$
h(x) = \frac{1}{n} \sum_{i=1}^{n}  f_i' (x) - \frac{1}{n} \sum_{i=1}^{n} f_i' (\hat{x_k}^{(i)}) = \frac{1}{n} \sum_{i=1}^{n} ( f_i' (x) - \frac{1}{n} \sum_{i=1}^{n}  f_i' (\hat{x_k}^{(i)}))
$$
It's obvious that there still exist $\hat{x_k}$ located in region of $\{ \hat{x_k}^{(1)}, \hat{x_k}^{(2)}, ..., \hat{x_k}^{(n)} \}$.

\textbf{Remark: } Although we give the proof of the average gradient in one dimension,  for the gradient of high-dimensional space,  the average gradient of high-dimensional space is the respective operation of each dimensional, so that it will be similar to one-dimensional method. This shows that the gradient of high-dimensional space also has the property of average gradient, which is similar to one-dimensional space, as in Corollary 1.1.

\textbf{Corollary 1.1} Assume $x \in \mathbb{R}^d$, where $d$ is the dimensions. And let assumption 1.1 hold. When N clients updates as the FedAvg, There still exist a $\hat{x_k}$ point, it satisfy that
\begin{align*}
    \nabla f(\hat{x_k}) = \frac{1}{N \tau} \sum_{i=1}^N \sum_{j=0}^{\tau - 1} \nabla f_i(x_{k,j}^{(i)})
\end{align*}

proof.
Consider $x \in \mathbb{R}^d$, we can assume that,
\begin{align*}
    x = [x_1, x_2, ... , x_d]
\end{align*}
it means that $x$ have d dimensions. 

Considering the multi-dimensional federated FedAvg process, we can treat each dimension as a FedAvg process. 
Then, according to Theorem 1.1, we know that there is a $\hat{x_i}$
point for the FedAvg process of $x_i$.
That means,
\begin{align*}
    \hat{x} = [\hat{x_1}, \hat{x_2}, ... , \hat{x_d}]
\end{align*}



\section{Convergence Analysis}
\label{appendix_convergence_analysis}

 
Here, we mainly give the proof of Theorem 2.1 and Theorem 3.1.

Among them, Theorem 2.1 need Lemma 2.1, 2.3, and 2.4.
Besides, we use lemma 1.2 here to analyze the Lighthouse convergence, which need lemma 2.1, 2.2, and 2.3.

\textbf{Difficulty Analysis:} The difficulty of convergence is that, the Lighthouse point $\hat{x_k}$ is exist, but cannot accurately get. Therefore, the proof of bound for Lighthouse $\hat{x_k}$ and normal $x_k$ is the key point. We provide some lemma about $\nabla f(x_k)^T \nabla f(\hat{x_k})$ bound can support the proof.

\subsection{Convergence of Lighthouse $\hat{x_k}$}

\textbf{Lemma 2.1} [Global one step]
\label{lemma1_new_theorem_1}
Let Assumption 2.1-2.2 hold. Consider $\alpha_k \leq \frac{1}{L}$, and $\alpha_k$ satisfy $\sum_1^{\infty} \alpha_k = + \infty$ and $\sum_1^{\infty} \alpha_k^2 < + \infty$, then we can get that 
$$
f(v_{k}) - f(x_k)  <=  - \alpha_k (1 - \frac{1}{2} L \alpha_k) \frac{\tau}{N} ||\nabla f(\hat{x_k}) ||^2 + \frac{1}{2} L \alpha_k^2 \sigma^2 \tau
$$


\textbf{proof.}
Considering the Federated process from $x_k$ to $v_k$, we can obtain that,
$$ 
\underbrace{f(v_{k}) - f(x_k)}_{T_2} = f(x_{k,\tau}) - f(x_{k,0}) = \sum_{j = 0}^{\tau - 1} \underbrace{ (f(x_{k,j+1}) - f(x_{k,j}))}_{T3}
$$
where $x_{k,j}$ represents a virtual aggregation point, which mainly draws on the work ~\cite{DBLP:conf/iclr/LiHYWZ20}.

For $T_3$,
$$f(x_{k,j+1}) - f(x_{k,j}) = \sum_{i = 1}^{N} p_i \underbrace{(f_i(x_{k,j+1}^{(i)}) - f_i(x_{k,j}^{(i)}))}_{T_4}
$$
For $T_4$, consider Lipschitz condition and Equ.\ref{lipschitz_condition1}, we get that
$$f_i(x_{k,j+1}^{(i)}) - f_i(x_{k,j}^{(i)}) \leq \nabla f_i(x_{k,j}^{(i)})^\mathrm{T} \underbrace{(x_{k,j+1}^{(i)} - x_{k,j}^{(i)})}_{T_5} + \frac{1}{2} L \underbrace{\Vert x_{k,j+1}^{(i)} - x_{k,j}^{(i)} \Vert^2}_{T_6}
$$
For $T_5$, we know that,
$$x_{k,j+1}^{(i)} = x_{k,j}^{(i)} - \alpha_k \nabla f_i(x_{k,j}^{(i)}, \zeta)
$$
For $T_6$, taking expectation for $\zeta$, we can obtain that
\begin{equation}
    \begin{aligned}
    \mathbb{E}_{\zeta} \Vert x_{k,j+1}^{(i)} - x_{k,j}^{(i)} \Vert^2 &= \alpha_k^2 \mathbb{E}_{\zeta} \Vert\nabla f_i(x_{k,j}^{(i)}, \zeta) - \nabla f_i(x_{k,j}^{(i)}) + \nabla f_i(x_{k,j}^{(i)}) \Vert^2 \\
    &= \alpha_k^2 (\sigma^2 + 2\mathbb{E}_{\zeta} (\nabla f_i(x_{k,j}^{(i))}, \zeta) - \nabla f_i(x_{k,j}^{(i)}))^\mathrm{T} \nabla f_i(x_{k,j}^{(i)}) + \mathbb{E}_{\zeta} \Vert\nabla f_i(x_{k,j}^{(i)} \Vert^2) \\
    &= \alpha_k^2 (\sigma^2 + \mathbb{E}_{\zeta} \Vert\nabla f_i(x_{k,j}^{(i)}) \Vert^2) \\
    \end{aligned}
\end{equation}
where the third equation is due to Assumption 2.

Then, for $T_4$, consider expectation for $\zeta$, and substitute above formulas, we obtain that
\begin{equation}
\label{T14}
    \begin{aligned}
    \mathbb{E}_{\zeta} f_i(x_{k,j+1}^{(i)}) - f_i(x_{k,j}^{(i)}) &\leq -\alpha_k \Vert \nabla f_i(x_{k,j}^{(i)}) \Vert^2 + \frac{1}{2} L \alpha_k^2 (\sigma^2 + \mathbb{E}_{\zeta} \Vert \nabla f_i(x_{k,j}^{(i)}) \Vert^2)\\
        &= - \alpha_k (1 - \frac{1}{2} L \alpha_k) \Vert \nabla f_i(x_{k,j}^{(i)})  \Vert^2 + \frac{1}{2} L \alpha_k^2 \sigma^2\\
    \end{aligned}
\end{equation}
Then, substitute Equ.\ref{T14} into $T_3$,
$$
\mathbb{E}_{\zeta} f(x_{k,j+1}) - f(x_{k,j}) \leq - \alpha_k (1 - \frac{1}{2} L \alpha_k) \underbrace{ \sum_{i=1}^{N} p_i \Vert\nabla f_i(x_{k,j}^{(i)}) \Vert^2}_{T_7} + \frac{1}{2} L \alpha_k^2 \sigma^2
$$
Considering $T_7$,
\begin{equation}
\begin{aligned}
\label{T16}
\sum_{i=1}^{N} p_i \Vert \nabla f_i(x_{k,j}^{(i)}) \Vert^2 &\geq \sum_{i=1}^{N} p_i^2 \Vert \nabla f_i(x_{k,j}^{(i)}) \Vert^2 \\
                &= \sum_{i=1}^{N} \Vert p_i \nabla f_i(x_{k,j}^{(i)}) \Vert^2 \\
                &\geq \frac{1}{N} \Vert \sum_{i=1}^{N} p_i \nabla f_i(x_{k,j}^{(i)})  \Vert^2 \\
                &= \frac{1}{N} \Vert \nabla f(x_{k,j}) \Vert^2 \\
\end{aligned}
\end{equation}
where the first inequality is due to $p_i \leq 1$, the second inequality is due to Cauchy inequality \ref{cauchy_inequality}.

Then, substitute Equ.\ref{T16} into $T_3$, we can obtain that
\begin{equation}
\label{T17}
    \mathbb{E}_{\zeta} f(x_{k,j+1}) - f(x_{k,j}) \leq - \alpha_k (1 - \frac{1}{2} L \alpha_k) \frac{1}{N} \Vert \nabla f(x_{k,j}) \Vert^2 + \frac{1}{2} L \alpha_k^2 \sigma^2
\end{equation}
And substitute Equ.\ref{T17} into $T_2$, we get that
\begin{equation}
\label{T18}
    \begin{aligned}
    \mathbb{E}_{\zeta} f(v_{k}) - f(x_k) &= \sum_{j=0}^{\tau - 1} \mathbb{E}_{\zeta} f(x_{k,j+1}) - f(x_{k,j}) \\
    &\leq  - \alpha_k (1 - \frac{1}{2} L \alpha_k) \frac{1}{N} \sum_{j=0}^{\tau - 1} \Vert\nabla f(x_{k,j}) \Vert^2 + \frac{1}{2} L \alpha_k^2 \sigma^2 \tau \\ 
    \end{aligned}
\end{equation}
Consider that
\begin{equation}
\label{T19}
    \begin{aligned}
    \Vert \nabla f(\hat{x_k}) \Vert ^2 &= \Vert \frac{1}{N \tau} \sum_{i=1}^N \sum_{j=0}^{\tau-1} \nabla f_i(x_{k,j}^{(i)}) \Vert^2 \\
    &= \Vert \frac{1}{\tau} \sum_{j=0}^{\tau-1} \nabla f(x_{k,j}) \Vert^2 \\
    &\leq \frac{1}{\tau} \sum_{j=0}^{\tau-1} \Vert \nabla f(x_{k,j})\Vert^2 \\
    \end{aligned}
\end{equation}
where the first equation is due to the definition of Lighthouse \ref{lighthouse}, the first inequality is due to Cauchy inequality \ref{cauchy_inequality}.

Then, substitute Equ.\ref{T19} into Equ.\ref{T18}, we can get that
$$
\mathbb{E}_{\zeta} f(v_{k}) - f(x_k)  \leq  - \alpha_k (1 - \frac{1}{2} L \alpha_k) \frac{\tau}{N} \Vert \nabla f(\hat{x_k}) \Vert^2 + \frac{1}{2} L \alpha_k^2 \sigma^2 \tau
$$

\textbf{Lemma 2.2} [Bound $\nabla f(x_k)^\mathrm{T} g(\hat{x_k})$] 
\label{lemma2_lemma1_new_theorem_1}
Let Assumption 2.1-2.3 hold, we can bound $\nabla f(x_k)^\mathrm{T} g(\hat{x_k})$ as follow:
$$
- \nabla f(x_k)^\mathrm{T} g(\hat{x_k}) \leq - (1 - \frac{1}{2} L \alpha_k) \frac{1}{N} \Vert \nabla f(\hat{x_k}) \Vert^2 - \frac{\mu \alpha_k \tau}{2} \Vert g(\hat{x_k}) \Vert^2  + \frac{1}{2} L \alpha_k \sigma^2 
$$

\textbf{proof.}
According to the $\mu$-strong convex Equ.\ref{strong_convex1}, we can get that
\begin{equation}
\label{T22}
    f(v_k) - f(x_k) \geq \nabla f(x_k)^\mathrm{T} (v_k - x_k) + \frac{\mu}{2} \Vert v_k - x_k \Vert^2
\end{equation}

Then, we consider that,
\begin{equation}
\label{T23}
    v_k = \frac{1}{N} \sum_{i=1}^{N} x_{k,\tau}^{(i)}, x_k = \sum_{i=1}^N x_{k,0}^{(i)}
\end{equation}

And, we know that,
\begin{equation}
\label{T24}
    x_{k,\tau}^{(i)} = x_{k,0} - \alpha_k \sum_{j=0}^{\tau - 1} \nabla f_i (x_{k,j}^{(i)}, \zeta)
\end{equation}

Therefore, combine Equ.\ref{T23} and Equ.\ref{T24}, we can obtain that
\begin{equation}
\begin{aligned}
\label{T25}
    v_k - x_k &= - \alpha_k \frac{1}{N} \sum_{i=1}^N \sum_{j=0}^{\tau-1} \nabla f_i (x_{k,j}^{(i)}, \zeta) \\
    &= -\alpha_k \tau g(\hat{x_k}) \\
\end{aligned}
\end{equation}
where the second equality is due to the definition of average gradient.

Then, substitute Equ.\ref{T25} into Equ.\ref{T22}, we can get that
\begin{equation}
\label{T26}
    \begin{aligned}
    f(v_k) - f(x_k) \geq - \alpha_k \tau \nabla f(x_k)^\mathrm{T} g(\hat{x_k}) + \frac{\mu \alpha_k^2 \tau^2}{2} \Vert g(\hat{x_k}) \Vert^2
    \end{aligned}
\end{equation}

According to the result of Lemma 1.1, we can bound Equ.\ref{T26} like below,
\begin{equation}
\label{T27}
\begin{split}
    - \alpha_k \tau \nabla f(x_k)^\mathrm{T} g(\hat{x_k}) &+ \frac{\mu \alpha_k^2 \tau^2}{2} \Vert g(\hat{x_k}) \Vert^2 \\
    &\leq f(v_k) - f(x_k) \leq - \alpha_k (1 - \frac{1}{2} L \alpha_k) \frac{\tau}{N} \Vert \nabla f(\hat{x_k}) \Vert^2 + \frac{1}{2} L \alpha_k^2 \sigma^2 \tau
\end{split}
\end{equation}

From Equ.\ref{T27}, we can obtain that
$$
- \nabla f(x_k)^\mathrm{T} g(\hat{x_k}) \leq - (1 - \frac{1}{2} L \alpha_k) \frac{1}{N} \Vert \nabla f(\hat{x_k}) \Vert^2 - \frac{\mu \alpha_k \tau}{2} \Vert g(\hat{x_k}) \Vert^2  + \frac{1}{2} L \alpha_k \sigma^2 
$$

\textbf{Lemma 2.3} (Enforce Positive Definiteness) Assume sequence ${\hat{B}_k}$ is generated by Hessian Updating process in our FedSSO algorithm. There exist constants 0 < $\underline{\kappa}$ < $\Bar{\kappa}$, such that $\{ \hat{B}_k^{-1}\}$ satisfies 
\begin{align*}
    \underline{\kappa} I \prec \hat{B}_k^{-1} \prec \Bar{\kappa} I
\end{align*}
where $I$ represent identity matrix. 

\textbf{proof.}
We mainly follow the techniques in \cite{2015AA},\cite{2016AA} for the analysis of bound of Hessian approximation, that is, by indirectly bounding the trace and determinant of $\hat{B}_k$, the eigenvalues of $\hat{B}_k$ is bounded.

In section 4.4, for the purpose of enforcing positive definiteness, we design Option 1 in Hessian Updating process on our FedSSO algorithm. For Option 1, we set $cur = \hat{y}_{k-1}^T s_{k-1}$. And,
\begin{align*}
    \lambda < \frac{\Vert \hat{y}_{k-1} \Vert^2}{cur} <\Lambda
\end{align*}
where $cur = \hat{y}_{k-1}^T s_{k-1}$ or $cur = \frac{2}{\lambda + \Lambda} \Vert \hat{y}_{k-1} \Vert^2$.

Now, we use induction method to prove the positive definiteness of $\hat{B}$.
Because of $\hat{B}_0 = I$, it satisfies positive definiteness.

Assume $\hat{B}_{k-1}$ also satisfies positive definiteness.

Let $Tr$ denote the trace of a matrix. Consider $k < R$, then we know that
\begin{align*}
    Tr(\hat{B}_k) &= Tr(\hat{B}_{k-1}) - \frac{\Vert \hat{B}_{k-1}s_{k-1} \Vert^2}{s_{k-1}^T \hat{B}_{k-1} s_{k-1}} + \frac{\Vert \hat{y}_{k-1} \Vert^2}{cur} \\
        &\leq Tr(\hat{B}_{k-1}) + \frac{\Vert \hat{y}_{k-1} \Vert^2}{cur} \\
        &\leq Tr(\hat{B}_{k-1}) + \Lambda \\
        &\leq Tr(\hat{B}_0) + k \Lambda \\
        &\leq Tr(\hat{B}_0) + k \Lambda \\
        &\leq M_3
\end{align*}
, for some constants $M_3$. Similarly, for iterations with $R$ as the cycle, the above bound will always hold.

Note that $k$ will take $R$ as the cycle, and the initial value of $\hat{B}$ will be reset to $I$. therefore, the above bound will be hold.

This implies that the largest eigenvalue of all matrices $\hat{B}$ is bounded uniformly.

Next, notice the fact that
\begin{align*}
    \hat{y}_{k-1}^T \hat{y}_{k-1} s_{k-1}^T s_{k-1} = \hat{y}_{k-1}^T s_{k-1} \hat{y}_{k-1}^T s_{k-1}
\end{align*}
by divide terms to left and right, we can get that
\begin{align*}
    \frac{ \hat{y}_{k-1}^T \hat{y}_{k-1}}{ \hat{y}_{k-1}^T s_{k-1}} = \frac{ \hat{y}_{k-1}^T s_{k-1}}{s_{k-1}^T s_{k-1}}
\end{align*}
According our setting on Hessian Update, the above formula means that,
\begin{align*}
    \frac{ \hat{y}_{k-1}^T \hat{y}_{k-1}}{cur} = \frac{cur}{s_{k-1}^T s_{k-1}}
\end{align*}
which means $\frac{cur}{s_{k-1}^T s_{k-1}}$ has the same bound as $\frac{ \hat{y}_{k-1}^T \hat{y}_{k-1}}{cur}$.

According to Powell ~\cite{1975Some}, we can derive an expression for the determinant of $\hat{B}_k$,
\begin{align*}
    det(\hat{B}_k) &= det(\hat{B}_{k-1}) \frac{cur}{s_{k-1}^\mathrm{T} \hat{B}_{k-1} s_{k-1}} \\
        &= det(\hat{B}_{k-1}) \frac{cur}{s_{k-1}^T s_{k-1}} \frac{s_{k-1}^T s_{k-1}}{s_{k-1}^\mathrm{T} \hat{B}_{k-1} s_{k-1}} \\
        &\geq det(\hat{B}_{k-1}) \lambda \frac{s_{k-1}^T s_{k-1}}{s_{k-1}^\mathrm{T} \hat{B}_{k-1} s_{k-1}} \\
        &\geq det(\hat{B}_{k-1})  \frac{\lambda}{M_3} \\
        &\geq det(\hat{B}_0) (\frac{\lambda}{M_3})^k \\
        &\geq M_4
\end{align*}
, for some constants $M_4$.

It shows the smallest eigenvalue of $\hat{B}_k$ is bounded away from zero.

Because the smallest and largest eigenvalues are both bound, it shows the bound of $\hat{B}_k$. It also shows there exist a bound for $\hat{B}_k^{-1}$ by easily quote a result from the literature \cite{2015AA}.

Next, We use lemma 1.2 here to analyze the Lighthouse convergence.

\textbf{Lemma 1.2} [Lighthouse convergence]
Let Assumptions 2.1-2.3 hold. $\underline{\kappa}$ and $\Bar{\kappa}$ are defined in Lemma 2.3. When $\alpha_k \le \frac{2}{L}$, $\eta_k \leq \frac{\mu \alpha_k}{2 L \Bar{\kappa}^2}$, and $\eta_k$ decays at the rate of $\mathcal{O} (\frac{1}{k})$, we can get that
\begin{multline*}
\mathbb{E}_{\zeta} f(x_{k+1}) - f(x_k) \leq - (1 - \frac{1}{2} L \alpha_k) \frac{\eta_k \underline{\kappa}}{N} \Vert\nabla f(\hat{x_k}) \Vert^2 \\- (\frac{\mu \alpha_k \tau \eta_k \underline{\kappa}}{2}
- {L \eta_k^2 \Bar{\kappa}^2} ) \Vert \nabla f(\hat{x_k}) \Vert^2 - \frac{\mu \alpha_k \tau \eta_k \underline{\kappa} \sigma^2}{2} +  \frac{1}{2} L \alpha_k \eta_k \underline{\kappa} \sigma^2 + {L \eta_k^2 \Bar{\kappa}^2} \sigma^2
\end{multline*}


\textbf{proof.}
Consider our FedSSO algorithm, we can know that
\begin{equation}
    x_{k+1} = x_k - \eta_k \hat{B}_k^{-1} g(\hat{x_k})
\end{equation}
which is consistent with the section B.4 for algorithm details.

According to the Lipschitz condition Equ.\ref{lipschitz_condition1}, we can get that
\begin{equation}
    \begin{aligned}
    f(x_{k+1}) - f(x_k) &\leq \nabla f(x_k)^\mathrm{T} (x_{k+1} - x_k) + \frac{L}{2} \Vert x_{k+1} - x_k \Vert^2\\
                        &= - \eta_k \nabla f(x_k)^\mathrm{T} \hat{B}_k^{-1} g(\hat{x_k}) + \frac{L \eta_k^2}{2} \Vert \hat{B}_k^{-1} g(\hat{x_k}) \Vert^2\\
                        &\leq - \eta_k \underline{\kappa} \nabla f(x_k)^\mathrm{T} g(\hat{x_k}) + \frac{L \eta_k^2 \Bar{\kappa}^2}{2} \Vert g(\hat{x_k}) \Vert^2\\
    \end{aligned}
\end{equation}

Then, substitute lemma 1.2 result into , we get that
\begin{equation}
\label{T37}
    \begin{aligned}
    f&(x_{k+1}) - f(x_k) \\
    &\leq - (1 - \frac{1}{2} L \alpha_k) \frac{\eta_k \underline{\kappa}}{N} \Vert\nabla f(\hat{x_k}) \Vert^2 - \frac{\mu \alpha_k \tau \eta_k \underline{\kappa}}{2} \Vert g(\hat{x_k}) \Vert^2 +  \frac{1}{2} L \alpha_k \eta_k \underline{\kappa} \sigma^2 + \frac{L \eta_k^2 \Bar{\kappa}^2}{2} \Vert g(\hat{x_k}) \Vert^2\\ 
                        &= - (1 - \frac{1}{2} L \alpha_k) \frac{\eta_k \underline{\kappa}}{N} \Vert\nabla f(\hat{x_k}) \Vert^2 - \frac{\eta_k}{2}(\mu \alpha_k \tau \underline{\kappa} - L \eta_k \Bar{\kappa}^2) \Vert g(\hat{x_k}) \Vert^2 + \frac{1}{2} L \alpha_k \eta_k \underline{\kappa} \sigma^2 \\
                        &= - A_1 \Vert\nabla f(\hat{x_k}) \Vert^2 - A_2 \Vert g(\hat{x_k}) \Vert^2 + A_3 \sigma^2
    \end{aligned}
\end{equation}
where $A_1 = (1 - \frac{1}{2} L \alpha_k) \frac{\eta_k \underline{\kappa}}{N}$, $A_2 = \frac{\eta_k}{2}(\mu \alpha_k \tau \underline{\kappa} - L \eta_k \Bar{\kappa}^2)$, $A_3 = \frac{1}{2} L \alpha_k \eta_k \underline{\kappa}$.

Taking expectation for $g(\hat{x_k})$, we can obtain that
\begin{equation}
\label{T38}
    \begin{aligned}
    \mathbb{E}_{\zeta} \Vert g(\hat{x_k}) \Vert^2 &= \mathbb{E}_{\zeta} \Vert g(\hat{x_k}) - \nabla f(\hat{x_k}) + \nabla f(\hat{x_k}) \Vert^2 \\
                                &= \mathbb{E}_{\zeta} \Vert g(\hat{x_k}) - \nabla f(\hat{x_k}) \Vert^2 + 2 \mathbb{E}_{\zeta} (g(\hat{x_k}) - \nabla f(\hat{x_k}))^\mathrm{T} \nabla f(\hat{x_k}) + \Vert \nabla f(\hat{x_k}) \Vert^2
    \end{aligned}
\end{equation}
Consider that
\begin{equation}
\label{T39}
    \begin{aligned}
    \mathbb{E}_{\zeta} \Vert g(\hat{x_k}) - \nabla f(\hat{x_k}) \Vert^2 &= \mathbb{E}_{\zeta} \Vert \frac{1}{N \tau} \sum_{i=1}^N \sum_{j=0}^{\tau - 1} \nabla f_i(x_{k,j}^{(i)}, \zeta) - \frac{1}{N \tau} \sum_{i=1}^N \sum_{j=0}^{\tau - 1} \nabla f_i(x_{k,j}^{(i)}) \Vert^2 \\
    &= \frac{1}{N^2 \tau^2} \mathbb{E}_{\zeta} \Vert  \sum_{i=1}^N \sum_{j=0}^{\tau - 1} ( \nabla f_i(x_{k,j}^{(i)}, \zeta) - \nabla f_i(x_{k,j}^{(i)})) \Vert^2 \\ 
    &\leq \frac{1}{N^2 \tau^2} N \tau \sum_{i=1}^N \sum_{j=0}^{\tau - 1} \mathbb{E}_{\zeta} \Vert  \nabla f_i(x_{k,j}^{(i)}, \zeta) - \nabla f_i(x_{k,j}^{(i)}) \Vert^2 \\
    &\leq \sigma^2
    \end{aligned}
\end{equation}
where the first equality is due to the definition of average gradient and Lighthouse, the first inequality is due to Cauchy inequality Equ.\ref{cauchy_inequality}, and the second inequality is due to our Assumption 2.

And we know that
\begin{equation}
\label{T40}
    \mathbb{E}_{\zeta}  g(\hat{x_k}) = \nabla f(\hat{x_k}) 
\end{equation}

Then, substitute Equ.\ref{T39} and Equ.\ref{T40} into Equ.\ref{T38}, we can obtain that
\begin{equation}
\label{T41}
    \mathbb{E}_{\zeta} \Vert g(\hat{x_k}) \Vert^2 \leq \sigma^2 + \Vert \nabla f(\hat{x_k}) \Vert^2
\end{equation}
Then, substitute Equ.\ref{T41} into Equ.\ref{T37}, we can obtain that
\begin{equation}
    \mathbb{E}_{\zeta} f(x_{k+1}) - f(x_k) \leq - (A_1+A_2) ||\nabla f(\hat{x_k}) ||^2 - A_2 \sigma^2 + A_3 \sigma^2 
\end{equation}


By accumulating the above formula, it can be seen that when $\alpha_k \le \frac{2}{L}$, $\eta_k \leq \frac{\mu \alpha_k}{2 L \Bar{\kappa}^2}$, and both $\alpha_k$ and $\eta_k$ decays at the rate of $\mathcal{O} (\frac{1}{k})$, $\hat{x_k}$ converges globally.


\subsection{Convergence of $x_k$}
Next, we consider the global convergence of $x_k$. In order to prove Theorem 2.1, we first introduce Lemma 2.4.

\textbf{Lemma 2.4} [Bound  $\nabla f(x_k)^T g(\hat{x_k})$]
Let Assumption 2.1-2.3 hold, then we can get that
\begin{align*}
    - \nabla f(x_k)^\mathrm{T} g(\hat{x_k}) \leq \frac{1}{\alpha_k \tau} [(1 - D)^{\tau} - 1] (f(x_{k}) - f^*) - \frac{\mu \alpha_k \tau}{2} \Vert g(\hat{x_k}) \Vert^2 + \frac{1}{2 \tau} L \alpha_k \sigma^2 \sum_{j=0}^{\tau} (1 - D)^j  
\end{align*}
where $D = \alpha_k (1 - \frac{1}{2} L \alpha_k) \frac{1}{N} \frac{2 \mu^2}{L}$, $f^*$ correspond to the optimal point $x^*$.

\textbf{proof.}
According to Assumption 1 and 3, we can get that
\begin{align*}
    || \nabla f(x_{k,j}) ||^2 &\geq \mu^2 \Vert x_{k,j} - x^* \Vert^2 \\
                              &\geq \frac{2 \mu^2}{L} (f(x_{k,j}) - f^*)  \tag{37}
\end{align*}
where the second inequality is due to L-smooth, and the first inequality is due to the $\mu$-strong convex.

Substitute the above Equ.37 into Equ.20, we can get that
\begin{align*}
    \mathbb{E}_{\zeta} f(x_{k,j+1}) - f(x_{k,j}) \leq - \alpha_k (1 - \frac{1}{2} L \alpha_k) \frac{1}{N} \frac{2 \mu^2}{L} (f(x_{k,j}) - f^*) + \frac{1}{2} L \alpha_k^2 \sigma^2
\end{align*}
According to the above formula, rearrange it,
\begin{align*}
    \mathbb{E}_{\zeta} f(x_{k,j+1}) - f^* \leq (1 - \alpha_k (1 - \frac{1}{2} L \alpha_k) \frac{1}{N} \frac{2 \mu^2}{L}) (f(x_{k,j}) - f^*) + \frac{1}{2} L \alpha_k^2 \sigma^2 \tag{38}
\end{align*}

Now, we consider iteration from $j=0$ to $j=\tau$. We can get that
\begin{align*}
    f(v_{k}) - f^* &= f(\sum_{i=1}^N p_i x_{k,\tau}^i) - f^* \\
                     &\leq \sum_{i=1}^N p_i f(x_{k,\tau}^i) - f^* \\
                     &= f(x_{k,\tau}) - f^* \\
                     &\leq (1 - D)^{\tau} (f(x_{k,0}) - f^*) + \sum_{j=0}^{\tau} (1 - D)^j \frac{1}{2} L \alpha_k^2 \sigma^2 \\
                     &= (1 - D)^{\tau} (f(x_{k}) - f^*) + \frac{1}{2} L \alpha_k^2 \sigma^2 \sum_{j=0}^{\tau} (1 - D)^j 
\end{align*} 
where the first inequality is due to convex function's property, the second inequality is according to Equ.38 and let $D = \alpha_k (1 - \frac{1}{2} L \alpha_k) \frac{1}{N} \frac{2 \mu^2}{L}$.

Then, we can get that
\begin{align*}
    f(v_{k}) - f(x_k) \leq [(1 - D)^{\tau} - 1] (f(x_{k}) - f^*) + \frac{1}{2} L \alpha_k^2 \sigma^2 \sum_{j=0}^{\tau} (1 - D)^j  
\end{align*}

Again, similar to Lemma 1.2 in Equ.28, we give a bound for $\nabla f(x_k)^T g(\hat{x_k})$, 
\begin{align*}
     - \alpha_k \tau \nabla f(x_k)^\mathrm{T} g(\hat{x_k}) &+ \frac{\mu \alpha_k^2 \tau^2}{2} \Vert g(\hat{x_k}) \Vert^2 \\
     &\leq f(v_k) - f(x_k) \leq [(1 - D)^{\tau} - 1] (f(x_{k}) - f^*) + \frac{1}{2} L \alpha_k^2 \sigma^2 \sum_{j=0}^{\tau} (1 - D)^j  
\end{align*}
Then, we can obtain that,
\begin{align*}
    - \nabla f(x_k)^\mathrm{T} g(\hat{x_k}) \leq \frac{1}{\alpha_k \tau} [(1 - D)^{\tau} - 1] (f(x_{k}) - f^*) - \frac{\mu \alpha_k \tau}{2} \Vert g(\hat{x_k}) \Vert^2 + \frac{1}{2 \tau} L \alpha_k \sigma^2 \sum_{j=0}^{\tau} (1 - D)^j  
\end{align*}

\textbf{Theorem 2.1} [Global Convergence]
Let Assumption 2.1-2.3 hold and $\beta, \gamma, u, \Gamma$ be defined therein. $\underline{\kappa}$, $\Bar{\kappa}$ be defined in Lemma 2.3. $f^*$ correspond to the optimal point $x^*$. Choose $\gamma^{-1} = \min \{ \frac{N L}{2 \underline{\kappa} \mu}, \frac{\mu}{2L} \}$, $\beta = \frac{2}{\mu}$, $\alpha_k = \eta_k \frac{L \Bar{\kappa}^2}{\mu \tau \underline{\kappa}} $ and $\eta_k = \frac{2}{\mu} \frac{1}{k + \gamma}$. Then, the FedSSO satisfies 
\begin{align*}
    \mathbb{E}[f(x_k)] - f^* \leq \frac{\nu}{k + \gamma}
\end{align*}
where $\nu = \max \{\frac{\beta^2 \Gamma}{\beta \mu - 1}, \frac{\mu}{2}(\gamma+1) \Omega_1 \}$, $\Omega_1 = \Vert x_1 - x^* \Vert^2$, and $\Gamma = \frac{L^2 \Bar{\kappa}^2 \sigma^2}{2 \mu \tau}$.

\textbf{proof.}
Substitute the above lemma 2.4 into Equ.36, we can get that,
\begin{align*}
    f&(x_{k+1}) - f(x_k) \\
    &\leq  \frac{\eta_k \underline{\kappa}}{\alpha_k \tau} [(1 - D)^{\tau} - 1] (f(x_{k}) - f^*) - \frac{\eta_k \underline{\kappa} \mu \alpha_k \tau}{2} \Vert g(\hat{x_k}) \Vert^2 + \frac{L \eta_k^2 \Bar{\kappa}^2}{2} \Vert g(\hat{x_k}) \Vert^2 + \frac{\eta_k \underline{\kappa}}{2 \tau} L \alpha_k \sigma^2 \sum_{j=0}^{\tau} (1 - D)^j \\
        &\leq - \frac{\eta_k \underline{\kappa}}{\alpha_k \tau} [1 - (1 - D)^{\tau}] (f(x_{k}) - f^*) - \frac{\eta_k}{2}(\mu \alpha_k \tau \underline{\kappa} - L \eta_k \Bar{\kappa}^2) \Vert g(\hat{x_k}) \Vert^2 + \frac{\eta_k \underline{\kappa}}{2 \tau} L \alpha_k \sigma^2 \tau  \tag{39}
\end{align*}
where the second inequality is due to the fact that $\sum_{j=0}^{\tau} (1 - D)^j \leq \tau$, and $0 < D < 1$.

For Equ.39, let $\psi = \frac{1}{\alpha_k \tau}[1 - (1 - D)^{\tau}]$, and $ \frac{\eta_k}{2}(\mu \alpha_k \tau \underline{\kappa} - L \eta_k \Bar{\kappa}^2) = 0$. We can get that
\begin{align*}
    f(x_{k+1}) - f^* &\leq (1 - \eta_k \underline{\kappa} \psi) (f(x_{k}) - f^*) + \frac{\eta_k \underline{\kappa}}{2 \tau} L \alpha_k \sigma^2 \tau \\
        &= (1 - \eta_k \underline{\kappa} \psi) (f(x_{k}) - f^*) + \eta_k^2 \frac{L^2 \Bar{\kappa}^2 \sigma^2}{2 \mu \tau} \tag{40}
\end{align*}
which means that $\alpha_k = \eta_k \frac{L \Bar{\kappa}^2}{\mu \tau \underline{\kappa}}$.

Now, we consider the bound of $\psi$ and $D$. Let $\alpha_k \leq \frac{1}{L}$, we can get that
$0 \leq D \leq \frac{\mu^2}{L^2 N} < 1$.

Notice the fact that $(1-D)^{\tau} \leq 1 - \tau D + \frac{\tau (\tau - 1)}{2} D^2$. Then, we can get that,
\begin{align*}
    \psi &\geq \frac{1}{\alpha_k \tau} [1-(1-\tau D + \frac{\tau (\tau - 1)}{2} D^2)] \\
      &= \frac{1}{\alpha_k \tau} [\tau D + \frac{\tau (\tau - 1)}{2} D^2)] \\
      &= \alpha_k^{-1} D (1+ \frac{\tau-1}{2}D) \\
      &= (1-\frac{1}{2}L\alpha_k)\frac{1}{N}\frac{2\mu^2}{L} (1 + \frac{\tau-1}{2}D) \\
      &\geq (1-\frac{1}{2}L\alpha_k)\frac{1}{N}\frac{2\mu^2}{L} \\
      &\geq \frac{\mu^2}{N L}
\end{align*}

Further, we know that
\begin{align*}
    1 - \eta_k \underline{\kappa} \psi \leq 1 - \eta_k \frac{\underline{\kappa} \mu^2}{N L}
\end{align*}

Therefore, for Equ.40, we can obtain that
\begin{align*}
    f(x_{k+1}) - f^* \leq (1 - \eta_k \frac{\underline{\kappa} \mu^2}{N L}) (f(x_{k}) - f^*) + \eta_k^2 \frac{L^2 \Bar{\kappa}^2 \sigma^2}{2 \mu \tau}
\end{align*}

Next, consider individually $\eta_k \leq \frac{N L}{\underline{\kappa} \mu^2}$, $\alpha_k \leq \frac{1}{L}$, $\alpha_k = \eta_k \frac{L \Bar{\kappa}^2}{\mu \tau \underline{\kappa}}$.
Let $\Delta_k = f(x_k) - f^*$, $u = \frac{\underline{\kappa} \mu^2}{N L}$, and $\Gamma = \frac{L^2 \Bar{\kappa}^2 \sigma^2}{2 \mu \tau}$, $\eta_k$ is a diminishing stepsize, we can know that,
\begin{align*}
    \Delta_{k+1} \leq (1-\eta_k u) \Delta_k + \eta_k^2 \Gamma \tag{41}
\end{align*}

Based on Equ.41, we learn from the work~\cite{DBLP:conf/iclr/LiHYWZ20}. Choose $\gamma^{-1} = \min \{ \frac{N L}{2 \underline{\kappa} \mu}, \frac{\mu}{2L} \}$, $\beta = \frac{2}{\mu}$, $\alpha_k = \eta_k \frac{L \Bar{\kappa}^2}{\mu \tau \underline{\kappa}} $, $\eta_k = \frac{\beta}{k + \gamma}$, $\nu = \max \{\frac{\beta^2 \Gamma}{\beta \mu - 1}, \frac{\mu}{2}(\gamma+1) \Omega_1 \}$, $\Omega_1 = \Vert x_1 - x^* \Vert^2$, and $\Gamma = \frac{L^2 \Bar{\kappa}^2 \sigma^2}{2 \mu \tau}$. We use induction to prove $\Delta_k \leq \frac{\nu}{k+\gamma}$. 

Firstly, the definition of $\nu$, $\beta$, and $\gamma^{-1}$ ensures that it holds for $k=1$. 
It's due to $\Delta_1 \leq \frac{\nu}{1 + \gamma}$ and $f(x_1) - f^* = \Delta_1 \geq \frac{\mu}{2} \Vert x_1 - x^* \Vert^2 = \frac{\mu}{2} \Omega_1$, which mean one bound of $\nu$. 
Also, $\eta_1 = \frac{\beta}{1 + \gamma} \leq \beta \gamma^{-1}$, and combine the bound for $\eta_k$ and $\alpha_k$, we can get the bound for $\gamma^{-1}$.

Assume conclusion holds for some $k$, it follows that 
\begin{align*}
    \Delta_{k+1} &\leq (1-\eta_k u) \Delta_k + \eta_k^2 \Gamma \\
        &\leq (1-\frac{\beta u}{k+\gamma}) \frac{\nu}{k+\gamma} + \frac{\beta^2 \Gamma}{(k+\gamma)^2} \\
        &= \frac{k+\gamma-1}{(k+\gamma)^2} \nu + \underbrace{[\frac{\beta^2 \Gamma}{(k+\gamma)^2} - \frac{\beta u - 1}{(k+\gamma)^2} \nu]}_{\leq 0} \\
        &\leq \frac{\nu}{k+\gamma+1}
\end{align*}
where the second inequality is by substituting $\eta_k$, the third inequality is based on the fact that $\frac{k-1}{k^2} \leq \frac{k-1}{k^2 - 1} = \frac{1}{k+1}$, and from the second equality, we can get the second bound for $\nu$.

Therefore, we can conclude that
\begin{align*}
    \mathbb{E}[f(x_k)] - f^* = \Delta_k \leq \frac{\nu}{k + \gamma}
\end{align*}
where $\nu = \max \{\frac{\beta^2 \Gamma}{\beta u - 1}, \frac{\mu}{2}(\gamma+1) \Omega_1 \}$, $\Omega_1 = \Vert x_1 - x^* \Vert^2$, and $\Gamma = \frac{L^2 \Bar{\kappa}^2 \sigma^2}{2 \mu \tau}$.

\subsection{Non Convex}
We consider the non convex condition. Before giving the theorem, we still give the lemma 3.1 about the lighthouse bound.


\textbf{Lemma 3.1} [Bound $- \nabla f(x_k)^\mathrm{T} \nabla f(\hat{x_k})$]
Let $\alpha \leq \frac{1}{2\sqrt{6} \tau L}$. Let Assumption 2.1,2.2 and 3.1 hold. It satisfies,
\begin{align*}
    - \nabla f(x_k)^\mathrm{T} \nabla f(\hat{x_k}) \leq - \frac{1}{2} (1-24\tau^2 \alpha^2 L^2) ||\nabla f(x_k)||^2 + 12\tau^2 \alpha^2 \sigma^2 L^2
\end{align*}

\textbf{proof.}

Let $N_1 = - \nabla f(x_k)^\mathrm{T} g(\hat{x_k})$:

\begin{align*}
    N_1 &= - <\nabla f(x_k), g(\hat{x_k}) - \nabla f(x_k) + \nabla f(x_k)> \\
        &= - || \nabla f(x_k) ||^2 + <\nabla f(x_k), \nabla f(x_k) - g(\hat{x_k}) > \\
        &\leq - || \nabla f(x_k) ||^2 + \frac{1}{2}|| \nabla f(x_k) ||^2 + \frac{1}{2} ||\nabla f(x_k) - g(\hat{x_k}) ||^2 \\
        &= - \frac{1}{2} || \nabla f(x_k) ||^2 + \frac{1}{2} ||\nabla f(x_k) - g(\hat{x_k}) ||^2 \\
\end{align*}





Bound $|| \nabla f(x_k) - \nabla f(\hat{x_k}) ||^2$. Here, we mainly consider the average weight $\frac{1}{N}$ for convenience. It can easily conduct to $p_i$.
\begin{align*}
    || \nabla f(x_k) - \nabla f(\hat{x_k}) ||^2 &= || \frac{1}{N \tau} \sum_{i=1}^{N} \sum_{j=0}^{\tau-1} \nabla f(x_{k,j}^i) - \frac{1}{N \tau} \sum_{i=1}^{N} \sum_{j=0}^{\tau-1} \nabla f(x_{k}) ||^2 \\
        &\leq \frac{1}{\tau} \sum_{j=0}^{\tau-1} || \nabla f(x_{k,j}) - \nabla f(x_k) ||^2 \\
        &\leq \frac{L^2}{\tau} \sum_{j=0}^{\tau-1} \underbrace{ || x_{k,j} - x_k ||^2}_{N_2} \\
\end{align*}

Consider $N_2$, 
\begin{align*}
    || x_{k,j} - x_k ||^2 &= || x_{k,j-1} - x_k - \alpha f(x_{k,j-1}, \zeta) ||^2 \\
        &\leq (1+\frac{1}{2\tau-1}) || x_{k,j-1} - x_k ||^2 + 2\tau ||\alpha f(x_{k,j-1}, \zeta) ||^2 \\
        &= (1+\frac{1}{2\tau-1}) || x_{k,j-1} - x_k ||^2 + 2\tau \alpha^2 || f(x_{k,j-1}, \zeta) - \nabla f(x_{k,j-1}) + f(x_{k,j-1}) - \nabla f(x_k) + \nabla f(x_k)||^2 \\
        &\leq (1+\frac{1}{2\tau-1}) || x_{k,j-1} - x_k ||^2 + 6\tau \alpha^2 \sigma^2 + 6\tau \alpha^2 ||f(x_{k,j-1}) - \nabla f(x_k)||^2 + 6\tau \alpha^2 ||\nabla f(x_k)||^2 \\
        &\leq (1+\frac{1}{2\tau-1} + 6\tau \alpha^2 L^2) || x_{k,j-1} - x_k ||^2 + 6\tau \alpha^2 \sigma^2 + 6\tau \alpha^2||\nabla f(x_k)||^2 \\
        &\leq (1+\frac{1}{\tau-1}) || x_{k,j-1} - x_k ||^2 + 6\tau \alpha^2 \sigma^2 + 6\tau \alpha^2||\nabla f(x_k)||^2 \\
\end{align*}
where $\alpha \leq \frac{1}{2\sqrt{3} \tau L}$.

Further, unrolling the above recursion, we obtain that,
\begin{align*}
    || x_{k,j} - x_k ||^2 &\leq \sum_{p=0}^{j-1} (1+\frac{1}{\tau-1})^p (6\tau \alpha^2 \sigma^2 + 6\tau \alpha^2||\nabla f(x_k)||^2) \\
        &\leq (\tau-1)[(1+\frac{1}{\tau-1})^\tau - 1] (6\tau \alpha^2 \sigma^2 + 6\tau \alpha^2||\nabla f(x_k)||^2) \\
        &\leq 24\tau^2 \alpha^2 \sigma^2 + 24\tau^2 \alpha^2 ||\nabla f(x_k)||^2\\
\end{align*}
where the third inequality is based on the fact that $(1+\frac{1}{\tau-1})^\tau \leq 5$.

Then, we can get that,
\begin{align*}
    || \nabla f(x_k) - \nabla f(\hat{x_k}) ||^2 \leq 24\tau^2 \alpha^2 \sigma^2 L^2 + 24\tau^2 \alpha^2 L^2 ||\nabla f(x_k)||^2
\end{align*}

Then, we can get that,
\begin{align*}
    N_1 &\leq - \frac{1}{2} || \nabla f(x_k) ||^2 + \frac{1}{2} ||\nabla f(x_k) - g(\hat{x_k}) ||^2\\
       &\leq - \frac{1}{2} || \nabla f(x_k) ||^2 + 12\tau^2 \alpha^2 \sigma^2 L^2 + 12\tau^2 \alpha^2 L^2 ||\nabla f(x_k)||^2 \\
       &= - \frac{1}{2} (1-24\tau^2 \alpha^2 L^2) ||\nabla f(x_k)||^2 + 12\tau^2 \alpha^2 \sigma^2 L^2
\end{align*}
where $\alpha <= \frac{1}{2\sqrt{6} \tau L}$.


\textbf{Theorem 3.1}[Non Convex] Let assumption non-convex conditions 2.1,2.2, and 3.1 hold. Let $\alpha_k = \frac{1}{2\sqrt{6} \tau L k}$ and $\eta_k = \frac{1}{\sqrt{k}}$. $\underline{\kappa}$ and $\Bar{\kappa}$ are defined in lemma 2.3. $\sigma$, $G$, and $L$ are defined on assumptions. we can conclude that
\begin{align*}
    \min_{1<k<K} ||\nabla f(x_k)||^2  &\leq \mathcal{O}(\frac{f(x_1) - f(x_K)}{\sqrt{K} \underline{\kappa} (1-\varTheta)}) + \mathcal{O} (\frac{\sigma^2}{K^2 (1-\varTheta)})\\ + \mathcal{O} (\frac{\Bar{\kappa}^2 G^2 L}{K \underline{\kappa} (1-\varTheta)})
\end{align*}
where $\varTheta = 24\tau^2 \alpha^2 L^2$.

proof.
Consider one global descent, we know that,
\begin{equation*}
    x_{k+1} = x_k - \eta \hat{B}_k^{-1} g(\hat{x_k})
\end{equation*}
According to the Lipschitz condition Equ.\ref{lipschitz_condition1}, we can get that
\begin{equation}
    \begin{aligned}
    f(x_{k+1}) - f(x_k) &\leq \nabla f(x_k)^\mathrm{T} (x_{k+1} - x_k) + \frac{L}{2} \Vert x_{k+1} - x_k \Vert^2\\
                        &= - \eta_k \nabla f(x_k)^\mathrm{T} \hat{B}_k^{-1} g(\hat{x_k}) + \frac{L \eta_k^2}{2} \Vert \hat{B}_k^{-1} g(\hat{x_k}) \Vert^2\\
                        &\leq - \eta_k \underline{\kappa} \nabla f(x_k)^\mathrm{T} g(\hat{x_k}) + \frac{L \eta_k^2 \Bar{\kappa}^2}{2} \Vert g(\hat{x_k}) \Vert^2\\
    \end{aligned}
\end{equation}

Substitute $N_1$ into above formula, we can get that
\begin{align*}
    f(x_{k+1}) - f(x_k) &\leq - \frac{\eta_k \underline{\kappa}}{2} (1-24\tau^2 \alpha^2 L^2) ||\nabla f(x_k)||^2 + 12\eta_k \underline{\kappa} \tau^2 \alpha^2 \sigma^2 L^2 + \frac{L \eta_k^2 \Bar{\kappa}^2}{2} \Vert g(\hat{x_k}) \Vert^2\\
\end{align*}

Next, we can discuss $|| g(\hat{x_k}) ||^2$ from two aspects.

First,
 According the bound for Gradient, we can get that,

\begin{align*}
    f(x_{k+1}) - f(x_k) &\leq - \frac{\eta_k \underline{\kappa}}{2} (1-24\tau^2 \alpha^2 L^2) ||\nabla f(x_k)||^2 + 12\eta_k \underline{\kappa} \tau^2 \alpha^2 \sigma^2 L^2 + \frac{L \eta_k^2 \Bar{\kappa}^2}{2} G^2\\
\end{align*}

Further, we can get that
\begin{align*}
    \sum_{k=1}^K \frac{\eta_k \underline{\kappa}}{2} (1-24\tau^2 \alpha^2 L^2) ||\nabla f(x_k)||^2 \leq f(x_1) - f(x_K) + \sum_{k=1}^K 12\eta_k \underline{\kappa} \tau^2 \alpha^2 \sigma^2 L^2 + \sum_{k=1}^K \frac{L \eta_k^2 \Bar{\kappa}^2}{2} G^2 
\end{align*}
That is,
\begin{align*}
    \frac{\eta_k \underline{\kappa} K}{2} (1-24\tau^2 \alpha^2 L^2) \min_{1<k<K} ||\nabla f(x_k)||^2 &\leq f(x_1) - f(x_K) + \sum_{k=1}^K 12\eta_k \underline{\kappa} \tau^2 \alpha^2 \sigma^2 L^2 + \sum_{k=1}^K \frac{L \eta_k^2 \Bar{\kappa}^2}{2} G^2 \\
        &\leq f(x_1) - f(x^*) + \sum_{k=1}^K 12\eta_k \underline{\kappa} \tau^2 \alpha^2 \sigma^2 L^2 + \sum_{k=1}^K \frac{L \eta_k^2 \Bar{\kappa}^2}{2} G^2 \\
\end{align*}
where the second inequality is based on the fact that $f(x_k) \ge f(x^*)$.

From the above inequality, we can get that
\begin{align*}
    \min_{1<k<K} ||\nabla f(x_k)||^2 &\leq \frac{2}{\eta_k \underline{\kappa} K (1-\varTheta)} (f(x_1) - f(x^*)) + \frac{24 \alpha^2 \tau^2 \sigma^2 L^2}{1-\varTheta} + \frac{\eta_k^2 \Bar{\kappa}^2 G^2 L}{\underline{\kappa} (1-\varTheta)}
\end{align*}
where let $\varTheta = 24\tau^2 \alpha^2 L^2$, and $\alpha \leq \frac{1}{2\sqrt{6} \tau L}$.

when $\alpha_k = \frac{1}{2\sqrt{6} \tau L k}$, $\eta_k = \frac{1}{\sqrt{k}}$, we can conclude that
\begin{align*}
    \min_{1<k<K} ||\nabla f(x_k)||^2  \leq \mathcal{O}(\frac{f(x_1) - f(x^*)}{\sqrt{K} \underline{\kappa} (1-\varTheta)}) + \mathcal{O} (\frac{\sigma^2}{K^2 (1-\varTheta)}) + \mathcal{O} (\frac{\Bar{\kappa}^2 G^2 L}{K \underline{\kappa} (1-\varTheta)})
\end{align*}
where $\varTheta = 24\tau^2 \alpha^2 L^2$.








\end{document}